\documentclass[journal]{IEEEtran}
\usepackage{array}
\usepackage{cite}
\usepackage{caption}
\usepackage{amsmath,graphicx}
\usepackage{epstopdf}
\usepackage{xcolor}
\usepackage{amssymb}
\usepackage{pifont}
\usepackage{booktabs}
\usepackage{array}
\usepackage{tabu}
\usepackage{booktabs}
\usepackage{graphicx}
\usepackage{subfigure}
\usepackage{bm}
\usepackage{multirow}
\usepackage{graphicx}
\usepackage{color}
\usepackage{float}
\usepackage{stfloats}
\usepackage[misc]{ifsym}
\usepackage{stmaryrd}
\usepackage{grffile}
\usepackage{makecell}
\usepackage{threeparttable}
\usepackage{bm}
\usepackage{enumitem}
\usepackage{enumerate}
\usepackage[pagebackref=true,breaklinks=true,letterpaper=true,colorlinks,bookmarks=false]{hyperref}
\providecommand{\tabularnewline}

\begin{document}\par

\title{MTU-Net: Multi-level TransUNet for Space-based Infrared Tiny Ship Detection}

\author{{Tianhao Wu, Boyang Li, Yihang Luo, Yingqian Wang, Chao Xiao, Ting Liu, Jungang Yang*, Wei An, Yulan Guo}

\thanks{This work was partially supported in part by the National Natural Science Foundation of China (Nos. 61972435, 61401474, 61921001, 62001478)}

\thanks{Tianhao Wu, Boyang Li, Yihang Luo, Yingqian Wang, Chao Xiao, Ting Liu, Jungang Yang, Wei An, Yulan Guo are with the College of Electronic Science and Technology, National University of Defense Technology (NUDT), P. R. China. Yulan Guo is also with the School of Electronics and Communication Engineering, Sun Yat-sen University, P. R. China. Emails: \{wutianhao16, liboyang20, luoyihang, wangyingqian16, xiaochao12, liuting, yangjungang, anwei, yulan.guo\}@nudt.edu.cn. (Corresponding author: Jungang Yang)}
}
\markboth{Journal of \LaTeX\ Class Files,~Vol.~14, No.~8, August~2015}%
{Shell \MakeLowercase{\textit{et al.}}: Bare Demo of IEEEtran.cls for IEEE Journals}
\maketitle
\begin{abstract}
Space-based infrared tiny ship detection aims at separating tiny ships from the images captured by earth orbiting satellites. Due to the extremely large image coverage area (e.g., thousands square kilometers), candidate targets in these images are much smaller, dimer, more changeable than those targets observed by aerial-based and land-based imaging devices. Existing short imaging distance-based infrared datasets and target detection methods cannot be well adopted to the space-based surveillance task. To address these problems, we develop a space-based infrared tiny ship detection dataset (namely, NUDT-SIRST-Sea) with $48$ space-based infrared images and $17598$ pixel-level tiny ship annotations. Each image covers about $10000$ square kilometers of area with $10000\times10000$ pixels. Considering the extreme characteristics (e.g., small, dim, changeable) of those tiny ships in such challenging scenes, we propose a multi-level TransUNet (MTU-Net) in this paper.  {Specifically, we design a Vision Transformer (ViT) Convolutional Neural Network (CNN) hybrid encoder to extract multi-level features}. Local feature maps are first extracted by several convolution layers and then fed into the multi-level feature extraction module (MVTM) to capture long-distance dependency. We further propose a copy-rotate-resize-paste (CRRP) data augmentation approach to accelerate the training phase, which effectively alleviates the issue of sample imbalance between targets and background. Besides, we design a FocalIoU loss to achieve both target localization and shape description. Experimental results on the NUDT-SIRST-Sea dataset show that our MTU-Net outperforms traditional and existing deep learning based SIRST methods in terms of probability of detection, false alarm rate and intersection over union. Our code is available at: \href{https://github.com/TianhaoWu16/Multi-level-TransUNet-for-Space-based-Infrared-Tiny-ship-Detection}{https://github.com/TianhaoWu16/Multi-level-TransUNet-for-Space-based-Infrared-Tiny-ship-Detection}.
\end{abstract}
\begin{figure}[t]
 \centering
 \includegraphics[width=8.65cm]{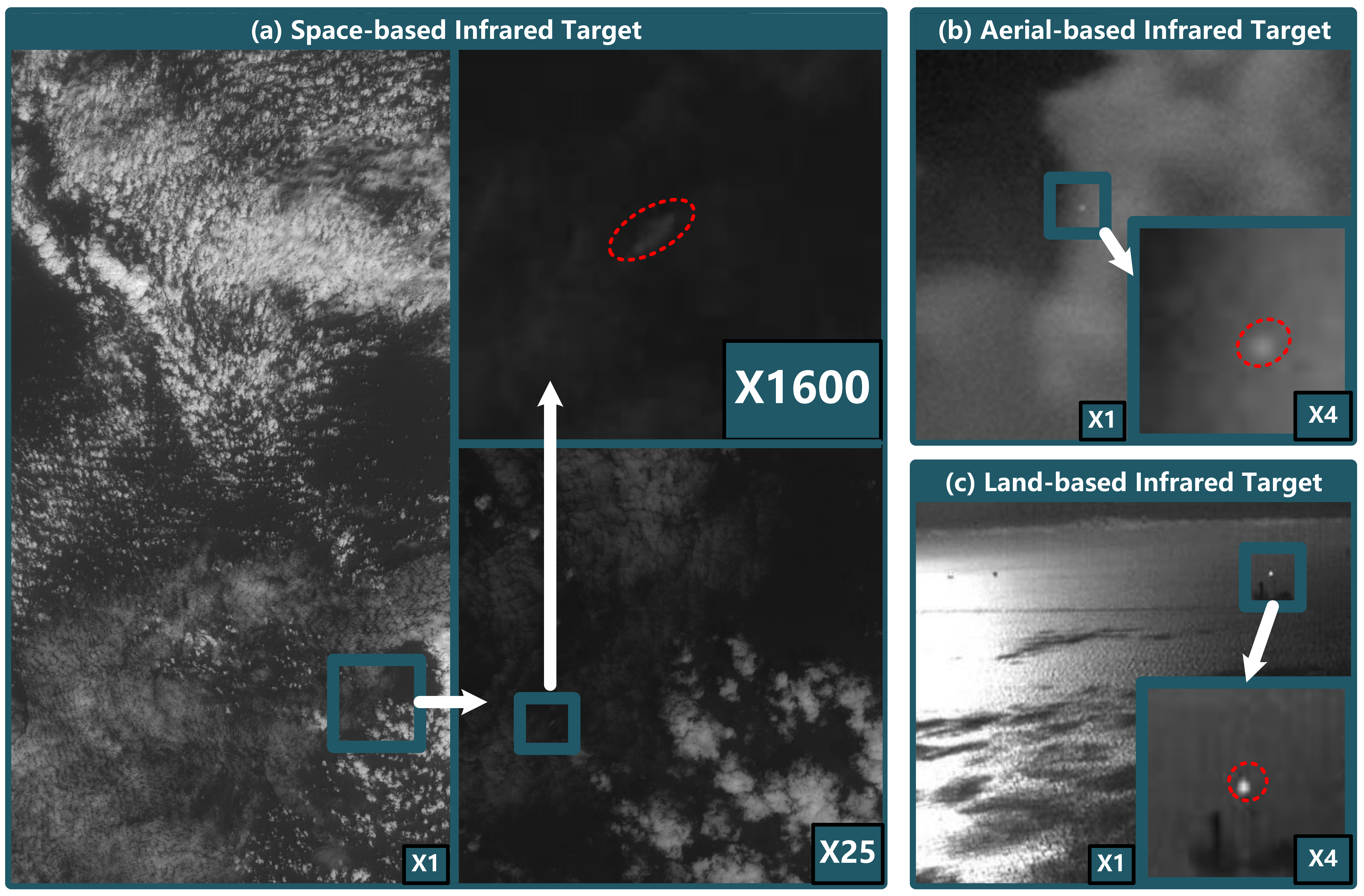}
 \caption{Comparison of space-based, air-based and land-based infrared images. (a) Typical image zoomed in $1$, $25$ and $1600$ times in our space-based SIRST dataset; (b) Typical image {zoomed in $1$ and $4$ times} in aerial-based SIRST dataset; (c) Typical image {zoomed in $1$ and $4$ times} in land-based SIRST. The targets are highlighted by \textcolor{red}{red} dotted circles respectively.
 }\label{fig1}
 \end{figure}
\section{Introduction}\label{introduction}
\IEEEPARstart{S}{pace-based} infrared tiny ship detection aims at separating tiny ships from the images captured by various (e.g., low, middle, geostationary) earth orbiting satellites \cite{1,2}. Due to the much longer imaging distance, the targets of space-based infrared images of ocean scenes exhibit several different characteristics \cite{1,2} (e.g., \textbf{larger image size}, \textbf{more complex background}, \textbf{more suspicious targets}, \textbf{extremely small targets}, \textbf{multi-scale targets}) from that of previous land-based and aerial-based single-frame infrared small target (SIRST) detection\cite{3,4,5}.

\par To detect tiny targets under complex scenes by infrared band, numerous traditional methods have been proposed, including filtering-based methods\cite{6,7}, local-contrast-based methods \cite{8,9,10,11,12,13}, and low-rank-based methods \cite{14,15,16,17,18,19,5}. Although promising progress has been achieved, these methods essentially rely on handcrafted features and fixed hyper-parameters. When the scenes (e.g., land, ocean, ports and clouds background) change dramatically, these methods suffer from a significant decrease in probability of detection ($P_d$) and an increase in false alarm rate ($F_a$).

\par {With the advances of deep learning, numerous Convolutional Neural Network (CNN) based methods \cite{20,21,22,23,24,25,26,27,28} have been proposed recently, introducing significant performance improvement in SIRST.} Dai et al. \cite{23} proposed the first segmentation-based SIRST detection network (ACM). Then, Dai et al. \cite{24} improved ACM by introducing a dilated local contrast measure and developed an ALC-Net. Moreover, Wang et al. \cite{25} used a conditional generative adversarial network (MFvsFA-cGAN) to achieve a trade-off between miss detection and false alarm for infrared small target detection. Li et al. \cite{27} proposed a dense nested attention network (DNA-Net) to extracted high-level information of small targets. However, the above CNN-based methods are designed for the short-distance imaging SIRST detection task (e.g., land-based and aerial-based SIRST). The candidate targets in the space-based tiny ship detection task are much smaller, dimer, and more changeable than those targets observed by aerial-based and land-based imaging devices. These methods cannot be well adopted to handle such challenges in space-based SIRST tiny ship detection.

\par To address the above problems, we first develop a space-based infrared tiny ship dataset, NUDT-SIRST-Sea. It contains $17598$ tiny annotated ships, and $48$ images with $10000\times10000$ pixels captured by cameras mounted on {the} low earth-orbiting satellite. {As shown in Fig. \ref{fig1}}, the tiny ships in space-based ocean scenes are visually non-salient in local regions compared to those targets in the land-based and aerial-based SIRST dataset. General CNN-based detection methods are not good at capturing long-distance dependency between targets and background. Therefore, it is necessary to further exploit the contextual relationship to achieve improved detection performance.
\begin{table*}[htb]
\setlength\tabcolsep{1.5pt}
\renewcommand\arraystretch{1.5}
 \centering
 \caption{Main characteristics of several popular SIRST datasets. Note that, our NUDT-SIRST-Sea dataset has images of the highest resolution, ground truth label type, lowest target to background ratio, largest target number, {lowest average SNR, smallest average target size} and the largest number of ground truth annotations compared with the mainstream SIRST datasets.}
  \scalebox{0.95}{
 \begin{tabular}{|l|c|c|c|c|c|c|c|c|c|}
 \hline
 Datasets & Image Type & Resolution & Label Type & Scene Type& {Target/Background ratio} & {Average SNR}& {Average Target size}&{Target number} \\ \hline
 NUAA-SIRST\cite{23} & real & $256\times256$ & Coarse Label & Aerial-based & $0.06097\% $ &{0.91}&{40} &$533$ \\\hline
 NUST-SIRST\cite{25} & synthetic & $320\times240$ & Coarse Label &Land-based & $0.94856\% $&{0.93}&{151}& $10337$ \\\hline
 IRSTD-1k\cite{28} & real & $256\times256$ & Ground Truth & Land-based & $0.02594\%$ &{0.76}&{85}& $1001$ \\\hline
 NUDT-SIRST\cite{27} & synthetic & $256\times256$ & Ground Truth & Aerial/Land-based & $0.06778\%$ &{0.88}&{44}& $1901$ \\\hline
 \textbf{NUDT-SIRST-Sea} & \textbf{real} &$\mathbf{10000\times10000}$& \textbf{ Ground Truth }& \textbf{ Space-based}& $\mathbf{0.000029\%}$ &\textbf{{0.28}}&\textbf{{29}}& $\mathbf{17598}$\\\hline
\end{tabular}}%
 \label{tab:datasts_compare}%
\end{table*}%

\par {Inspired by the success of the Vision Transformer \cite{29} (ViT) structure in generic object detection, we first design a multi-level ViT CNN hybrid encoder.} Specifically, we design a multi-level ViT module (MVTM) to achieve coarse-to-fine feature extraction. In our multi-level ViT CNN hybrid encoder, multi-level features are first extracted by CNN. Then, these features are refined by MVTM to capture long-distance dependency. Due to the sparsity of tiny ships in space-based infrared images, the foreground targets and background are extremely imbalanced. To address this issue, we propose a novel copy-rotate-resize-paste (CRRP) data augmentation approach to increase the ration of candidate targets in the training phase and ultimately accelerate the convergence of the network. Moreover, we find that existing IoU-like loss overly focuses on producing complete shape of the target, but lacks the ability to locate small-scale targets. The Focal loss focuses on hard samples but lacks of ability to produce complete shape of the target. Therefore, we design a novel FocalIoU loss to accurately localize tiny targets and completely produce the shape of the target. 

\par To the best of our knowledge, this is the first deep learning based work to achieve space-based infrared tiny ship detection. The contributions of our work can be summarized as follows:
\begin{itemize}
\item To our knowledge, NUDT-SIRST-Sea is the largest manually annotated dataset with a wide variety of categories in space-based infrared observation. $17598$ high-precision bounding boxes and pixel-level annotations are introduced to support the development and evaluation of various target detectors in space-based infrared images.
\item {We propose a novel Transformer CNN hybrid architecture (i.e., MTU-Net) for space-based infrared tiny ship detection. With the help {of} multi-level ViT CNN hybrid encoder, the long-distance dependency of tiny ships can be well incorporated and fully exploited by coarse-to-fine feature extraction and multi-level feature fusion.}
\item {A CRRP data augmentation method and a FocalIoU loss are proposed to alleviate the foreground-background imbalance problem and achieve `double-win' of target localization and shape description.}
\item Experimental results show that space-based infrared tiny ship detection is a challenging task, previous land-based and aerial-based SIRST methods can not well handle those challenges (e.g., extremely small and dim targets) introduced by this task. 
Our method can achieve state-of-the-art results in three metrics: probability of detection ($P_d$), false alarm rate ($F_a$), and intersection over union ($IoU$).
\end{itemize}

\par This paper is organized as follows: In Section \ref{SecMotivations}, we briefly describe the importance of our task and present the statistical characteristics and challenges of our NUDT-SIRST-Sea dataset in details. In Section \ref{SecRelatedWork}, we briefly review the related work. In Section \ref{SecMethodology}, we introduce the architecture of our MTU-Net, CRRP data augmentation approach, and our FocalIoU loss in details. The experimental results are represented in Section \ref{SecExperiment}. Section \ref{SecConclusion} gives the conclusion.

\begin{figure*}[t]
 \centering
 \includegraphics[width=18cm]{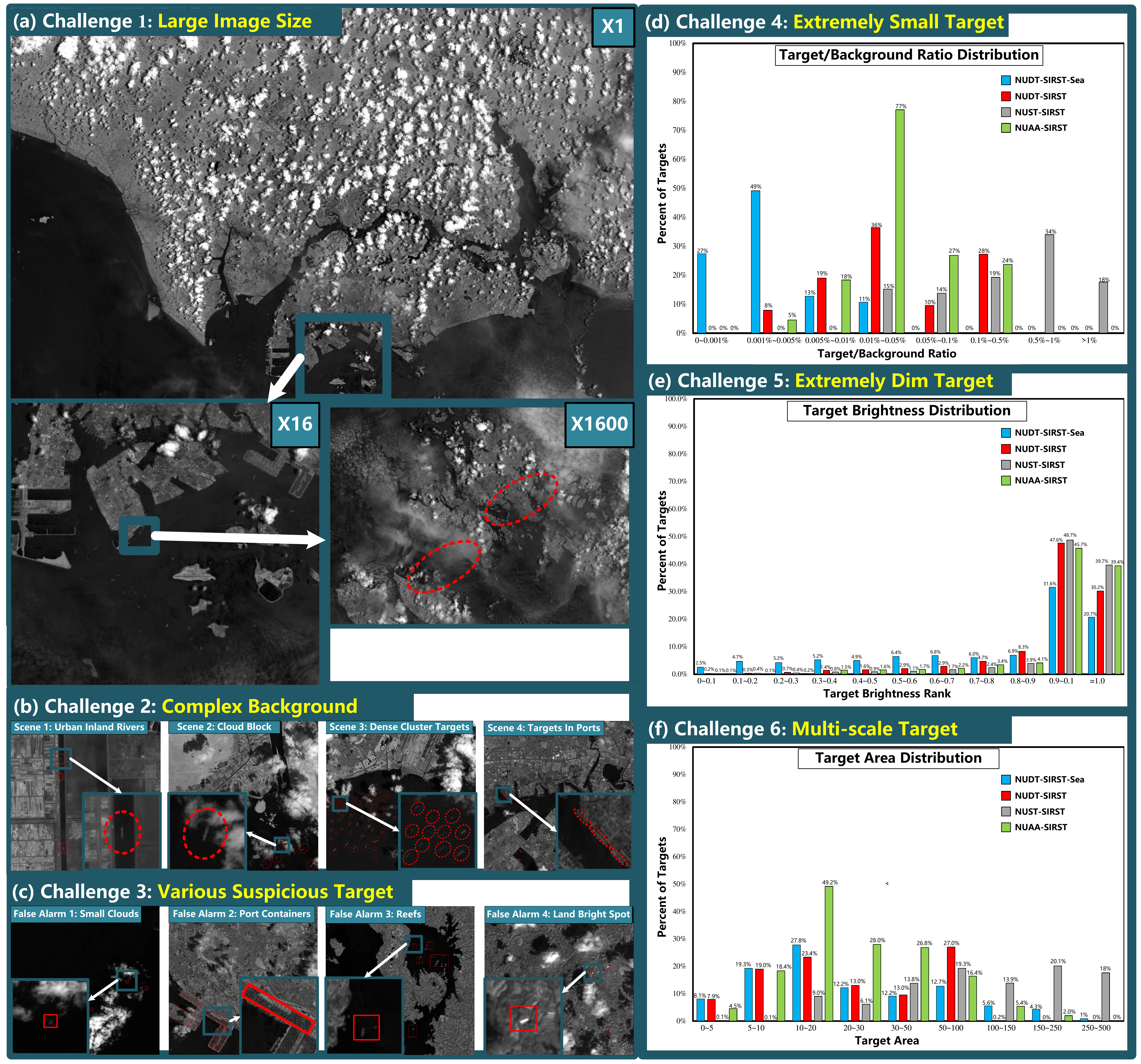}
 \setlength{\belowcaptionskip}{-0.2cm}
 \caption{Overall description of the NUDT-SIRST-Sea dataset; (a) Typical image zoomed in $1$, $16$ and $1600$ times with tiny ships in the NUDT-SIRST-Sea dataset; (b) Illustration of the variety of background; (c) Illustration of the variety of suspicious targets; (d) Distribution of target number with the target to background ratio; (e) Distribution of target number with target brightness; (f) Distribution of target number with target area.
 }\label{Cover}
 \end{figure*}
\section{{Challenges}}\label{SecMotivations}
\subsection{Importance}
Space-based infrared tiny ship detection is {one of the most important tasks} in the SIRST detection family, which generally {includes} land-based \cite{25}, aerial-based \cite{23,28}, and space-based SIRST detection tasks. As shown in Table \ref{tab:datasts_compare}, a space-based infrared image usually covers about $10000$ square kilometers of area with $10000\times10000$ pixels, thousands of times larger than the size of other land-based and aerial-based images. Due to the much longer imaging distance, the targets of space-based infrared images exhibit extreme characteristics (e.g., small, dim, changeable) than other SIRST images. Due to the lack of sufficient high-quality annotated datasets, existing deep learning based methods cannot work well on the space-based long-distance detection tasks. To alleviate this problem, we propose a space-based SIRST dataset (namely, NUDT-SIRST-Sea). {Specifically, We collect $48$ real images captured by sensors mounted on the low earth-orbiting satellite from both near infrared (i.e., 845 nm$\sim$885 nm wave length) and short infrared wave band (i.e., 1650 nm$\sim$1660 nm wave length). The field of view for both wave lengths images is completely overlapped. $41$ images are used for training and the rest $7$ images are used for testing.} Each image covers about 10000 square kilometers of area with $10000\times10000$ pixels. Moreover, we takes more than $500$ hours to manually generate $17598$ pixel-level tiny ship annotations. Both bounding boxes and pixel-level annotations are provided for the development and evaluation of various target detectors. {The average target size of our dataset is 29 pixels, which is much smaller than the size of those images in other datasets.} The target to background ratio of our NUDT-SIRST-Sea dataset is $0.000029\%$, which is hundreds of times smaller than the target to background ratios of other SIRST datasets.

\vspace{-1em}
\subsection{{Statistical Properties of NUDT-SIRST-Sea}}
\subsubsection{\textbf{Much Larger Image Size}}
compared with existing SIRST datasets in Table \ref{TabOVER_ALL}, each image of NUDT-SIRST-Sea covers about $10000$ square kilometers of area with $10000\times10000$ pixels, thousands of times larger than the image sizes of NUDT-SIRST \cite{27}, NUST-SIRST \cite{25} and NUAA-SIRST \cite{23}. As shown in Fig. \ref{Cover} (a), a much larger image contains more different scenes (e.g., port, land, clouds, sea). Besides, a much larger image size results in higher computational difficulties.

\subsubsection{\textbf{Much More Complex Background}}
as shown in Fig. \ref{fig1}, aerial-based and land-based infrared images are much simpler than space-based infrared images due to the limited coverage area. As shown in Fig. \ref{Cover} (b), different scenes (e.g., clouds, tiny ships, port, land, sea face) can form more types of complex scenes. Several scenes are considered as the difficult targets in NUDT-SIRST-Sea: urban inland river, cloud blocks, dense cluster targets, and targets in port. These complex scenes challenge the method's ability to capture long-distance context information.

\subsubsection{\textbf{Multi-Type Suspicious Targets}}
Figure \ref{Cover} (c) shows that our NUDT-SIRST (sea) dataset has a rich variety of suspicious targets, including tiny clouds, port containers, reefs, and land bright spots. These suspicious targets are very easily confused with real ship targets in shape and brightness and thus cause false alarm.

\subsubsection{\textbf{Much Smaller Targets}}
as shown in Table \ref{tab:datasts_compare}, {the average target size of our NUDT-SIRST-Sea dataset is only 29 pixels which is much smaller than the average target size of images in other mainstream SIRST datasets.} The target to background ratio of our NUDT-SIRST-Sea dataset is $0.000029\%$, hundreds of times smaller than the target to background ratios of NUDT-SIRST \cite{27}, NUST-SIRST \cite{25} and NUAA-SIRST \cite{23}. As shown in Fig. \ref{Cover} (d), $76\%$ targets cover less than $0.005\%$ area in space-based images. Targets of other datasets \cite{27, 25, 23, 28} mostly cover over $0.05\%$ area in space-based images. Therefore, much smaller targets in NUDT-SIRST-Sea make this dataset more challenging than other datasets.
\subsubsection{\textbf{Much Dimmer Targets}}
{as shown in Table \ref{tab:datasts_compare}, our NUDT-SIRST-Sea has the much smaller average target SNR than other datasets \cite{23,25,27,28}.} Detailed comparisons among these existing datasets are shown in Fig. \ref{Cover} (e). Datasets like NUDT-SIRST \cite{27}, NUST-SIRST \cite{25} and NUAA-SIRST \cite{23} mostly focus on bright targets. However, more than $20\%$ of targets have a brightness smaller than $0.5$ in our NUDT-SIRST-Sea. In contrast, less than $5\%$ of targets have a brightness smaller than $0.5$ in other aerial-based and land-based datasets. Compared to other datasets, NUDT-SIRST-Sea is more challenging on dim targets.

\subsubsection{\textbf{Multi-Scale Targets}}
as shown in Fig. \ref{Cover} (f), the size of different types of ships (e.g., large cruise ships, medium-sized oil recovery wells, small yachts) varies a lot, ranging from $2$ pixels to $500$ pixels. Due to the large area occupied by space-based infrared images, targets with different scales often appear in the same scene. Detecting targets with different scales in the same scene is a fairly challenging task.

\section{Related Work}\label{SecRelatedWork}
In this section, we briefly review the major works in {space-based visible tiny Ship Detection}, SIRST detection and Vision Transformer.
\subsection{{Space-based Visible Tiny Ship Detection}}
{Space-based visible tiny ship detection aims to detect tiny ship in remote sensing visible images. Chen et al. \cite{30} proposed a degraded reconstruction enhancement-based method for tiny ship detection in remote sensing images. They incorporated a CRoss-stage Multi-head Attention module in the detector to further improve the feature discrimination by leveraging the self-attention mechanism and a large-scale dataset was proposed in their work. Wu et al. \cite{31} proposed an effective Tiny Ship Detector for Low-Resolution RSIs, abbreviated as LR-TSDet. Their LR-TSDet was consisting of three key components: a filtered feature aggregation (FFA) module, a hierarchical-atrous spatial pyramid (HASP) module, and an IoU-Joint loss. Further, they introduced a new dataset called GF1-LRSD collected from the Gaofen–1 satellite for tiny ship detection in low-resolution RSIs. Li et al. \cite{32} proposed a new SDVI algorithm, named enhanced YOLO v3 tiny network for real-time ship detection. The algorithm can be used in video surveillance to realize the accurate classification and positioning of six types of ships (including ore carrier, bulk cargo carrier, general cargo ship, container ship, fishing boat, and passenger ship) in real-time.}

{However, the above methods are designed for visible tiny ship detection task. Tiny ships in infrared band are much dimmer and shapeless than those in RGB bands. Moreover, infrared images contain poorer contextual relation than visible images. Multi-type suspicious targets (i.e., tiny clouds, port containers, reefs and land bright spots) exhibit more texture and color differences from the tiny ships in visible images while are more easily confused with targets (i.e., ships under clouds, ships in port, ships by reefs) in infrared images.}
\subsection{Single-frame Infrared Small Target Detection}
General SIRST detection tasks (e.g., aerial-based and land-based SIRST) {have} been extensively investigated for decades. 
{Many filter-based background  methods \cite{6,7} were proposed. These methods used special designed filters for the background noise and clutter suppression. Considering that the small target is more visually salient than its surrounding background, human visual system local contrast based methods \cite{8,9,11,10,12,13} have been proposed. However, clutters can be are easily confused with targets in highlight scenes. To solve this problem, low rank based methods were proposed \cite{14,15,16,17,18,19,5}. Non-local self-correlation between background patches were used in infrared images to construct low-rank sparse decomposition model. Nevertheless, these previous filter-based methods, local contrast based methods and low rank based methods rely on fixed hyper-parameters. When real scenes change dramatically, such as in clutter background, target shape, and target size, it is difficult to use fixed hyper-parameters to handle such variations.}
\par Different from traditional methods, CNN-based methods adopt a data-driven training manner to learn common characteristics among small targets. Thanks to the previous open-sourced SIRST datasets and the powerful detection models, CNN-based methods have achieved promising progress recently. Dai et al. \cite{23} proposed the first segmentation-based CNN network. They designed an asymmetric contextual module to aggregate features from shallow layers and deep layers. Then, Dai et al. \cite{24} further improved their ACM by introducing a dilated local contrast measure in their ALC-Net. {Specifically, a feature cyclic shift scheme was designed to achieve a trainable local contrast measure.} After that, {Wang et al. \cite{25} decomposed the infrared target detection problem into two opposed sub-problems (i.e., miss detection and false alarm). They proposed a conditional generative adversarial network (MDvsFA) to achieve the trade-off between miss detection and false alarm for infrared small target detection.} {Considering pooling layers in the networks could lead to the loss of targets in deep layers,} Li et al.\cite{27} proposed a dense nested attention network (namely, DNA-Net). {With the help of their specifically-designed dense nested interactive module (DNIM), high-level information of small targets can be extracted and the response of small targets can also be maintained in the deep CNN layers.} {Zhang et al. \cite{28} proposed an infrared shape network (28). In their network, a Taylor finite difference (TFD)-inspired edge block and two-orientation attention aggregation (TOAA) block were devised to address the problem of submerging of infrared targets in the background of heavy noise and clutter.}
\par {Benefited from these specially-designed architectures and modules, the above CNN-based methods have achieved promising results in the land-based and aerial-based SIRST detection tasks. However, space-based SIRST images are quite different from the aerial-based and land-based SIRST ones. Smaller, dimer, and more changeable targets make it difficult to achieve high-performance detection under limited receptive fields introduced by CNN architecture.} Moreover, poor long-distance dependency capture ability of traditional CNN architecture may result in more false alarm. Therefore, it is necessary to introduce more long-distance information capture modules to further exploit the correlation of targets and background.

\subsection{Vision Transformer}
Inspired by the success of the Transformer architectures in the NLP area \cite{33}, some works try to apply them in the computer vision area. Vision Transformer \cite{29} (ViT) is the first work to apply Transformer to computer vision. It uses a non-overlapping medium-sized image patches in Transformer to achieve high precision image classification. With the success of ViT, more promising ViT works emerged. Various ViT-based structures are proposed to handle various high-level tasks (e.g., object detection, semantic segmentation). For example, Carion et.al \cite{34} proposed the first end-to-end object detection network with Transformers (namely, DETR). After that, Chen et al.\cite{35} proposed TransUNet and argued that Transformers can serve as powerful encoders for the medical image segmentation tasks with the combination of U-Net \cite{36} to enhance finer details by recovering localized spatial information. In SIRST detection field, Liu et al. \cite{37} proposed the first work to explore the Vision Transformer to detect infrared small-dim targets and achieved promising performance in SIRST. {They first used CNN to extract local features. Then, they adopted the ViT to high-level information of target localization from local features.} 

{Although achieving promising performance,  the above transformer-based works are not designed for spaced-based SIRST detection tasks. Specifically, DETR is designed for normal scale object detection and cannot well capture the features of tiny targets. TransUNet mainly focuses on the whole image segmentation performance and but pay less attention to local information of tiny targets. The ViT for SIRST method proposed by Liu et al. \cite{37} is mainly designed for aerial-based and land-based SIRST. However, space-based SIRST detection task requires both high level information for target localization and low level information for shape description. Their single-level ViT structure only applied on the features extracted by the last CNN layer. So their method cannot fully capture low level information for shape description and easily confuse the real targets with suspicious targets (e.g. tiny clouds, port containers, reefs, and land bright spots).} To address above problems, our MTU-Net combines multi-level ViT module (MVTM) and CNN in a multi-level ViT CNN hybrid encoder. CNN extracts multi-level features. Then, MVTM refines features to capture long-distance dependency of multi-level features.

\section{Methodology}\label{SecMethodology}
In this section, we introduce our multi-level TransUNet (Sections \ref{SecOverall}, \ref{Vit_CNN_encoder}, \ref{U-shape_decoder}, \ref{Eight_Module}), CRRP data augmentation method (Section \ref{Data Augmentation}), and FocalIoU loss (Section \ref{FocalIoU loss}) in details.
\begin{figure*}[t]
\centering
\setlength{\belowcaptionskip}{-0.5cm}
\includegraphics[width=18.2cm]{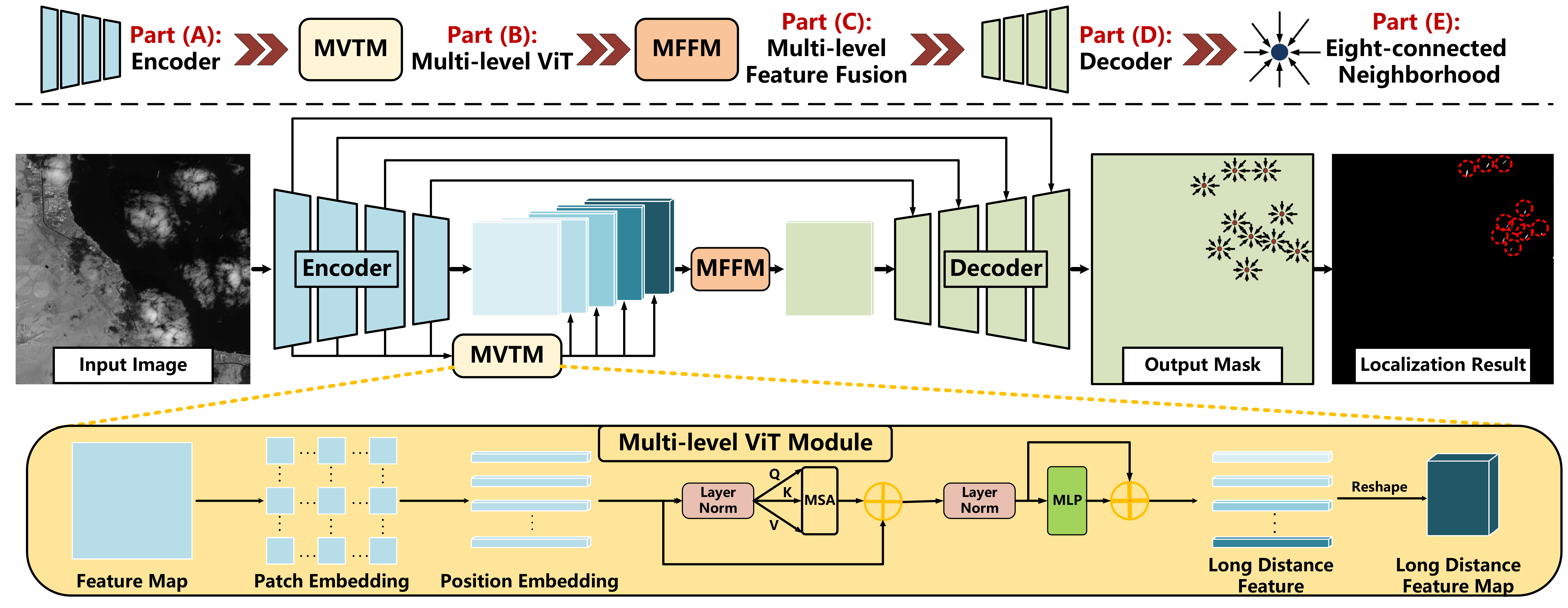}
\caption{An illustration of the proposed multi-level TransUNet in this paper.
(a) Encoder. Input image is fed into the CNN encoder to coarsely extract multi-scale features. (b) Multi-level ViT module (MVTM). Then, features of different levels go through the multi-level ViT module (MVTM) to extract long-distance features. (c) Multi-level feature fusion module (MFFM). The multi-level {features} are fed into the multi-level feature fusion module, where these features are concatenated and fused to incorporate long-distance information. (d) Decoder. Features with multi-level long-distance information are fed into the U-shape decoder and fused at the nodes of skip connection to generate final predicted probability map. (e) Eightconnected neighborhood clustering module. The predicted probability map is clustered and the centroid of each target region is finally determined.}\label{fig:MTU_overall}
\end{figure*}
\subsection{Overall Architecture}\label{SecOverall}
As illustrated in Fig. \ref{fig:MTU_overall}, our MTU-Net takes a single image as its input and sequentially consists of a multi-level ViT CNN hybrid encoder (Section \ref{Vit_CNN_encoder}), a U-shape decoder (Section \ref{U-shape_decoder}), and an eight-connected neighborhood clustering module (Section \ref{Eight_Module}) to generate the pixel-level localization and classification results.
\par Section \ref{Vit_CNN_encoder} introduces our multi-level ViT CNN hybrid encoder. {Input image is first cut into image patches $\bm{I}\in{\mathbb{R}}^{{C}\times{H}\times{W}}$, where $C$, $H$ and $W$ denote the channel, width and height of image patch. Image patches $\bm{I}$ are preprocessed before fed into the CNN to coarsely extract multi-scale features $\bm{F_i}\in{{\mathbb{R}}^{{C_i}\times{H_i}\times{W_i}}}(i\in{\{1, 2, \dots, k\}})$, where $k$ denotes the level numbers of CNN.}
Then, each feature of different levels ${F_i}(i\in{\{1, 2, \dots, k-1\}})$ goes through the multi-level ViT module (MVTM) to obtain ${\bm{V}_i}\in{{\mathbb{R}}^{{C_i}\times{H_k}\times{W_k}}}(i\in{\{1, 2, \dots, k-1\}})$.
The multi-level features $\{\bm{{V}_i}\}(i\in{\{1, 2, \dots, k-1\}}$ and $\bm{F_k}$ are fed into the multi-level feature fusion module (MFFM). In MFFM, features are concatenated and fed to a $1\times{1}$ convolution to generate the feature $\bm{{M}_k}\in{{\mathbb{R}}^{{C_k}\times{H_k}\times{W_k}}}$ with multi-level long-distance information. Section \ref{U-shape_decoder} introduces the U-shape decoder. Features $\bm{M_k}$ with multi-level long-distance information are fed into the decoder and fused with $\bm{F_i}(i\in{\{k-1, k-2, \dots, 1\}})$ at the nodes of skip connection to generate $\bm{M_i}(i\in{\{k-1, k-2, \dots, 1\}})$ and final predicted probability map $\bm{P}$. Section \ref{Eight_Module} elaborates the eight-connected neighborhood clustering module. {Final predicted probability maps} $\bm{P}$ are fed into this module to calculate the spatial locations of target centroid, which {are} then used for comparison in Section \ref{Comparison}. 
\vspace{-1em}

\subsection{Multi-level ViT CNN Hybrid Encoder}\label{Vit_CNN_encoder}
\subsubsection{{Motivation}}
{As shown in Fig. \ref{fig:MTU_overall}, our MTU-Net consists of a multi-level ViT CNN hybrid encoder, a U-shape decoder, and an eight-connected neighborhood clustering module to generate the pixel-level localization and classification results. To achieve efficient feature extraction for extremely large images (e.g., 10000x10000 resolution), the images are first cut into $1024\times1024$ patches and then fed into the ResNet-18 \cite{38} to extract multi-scale local features. To distinguish multi-type suspicious targets in complex backgrounds, more long-distance information is required. The proposed multi-level ViT module (MVTM) refines the extracted multi-scale local features. In this way, the long-distance dependency of suspicious targets in complex background are captured from high-level features. Moreover, multi-scale infrared small targets are significantly different in their sizes, ranging from one pixel (i.e., point targets) to tens of pixels (i.e., extended targets). With the increase of network layers, high-level information of target localization is obtained, while the shape description of extended targets is easily lost after multiple down-samplings. Therefore, we designed a multi-level feature fusion module (MFFM) to fuse multi-level features extracted by MVTM. In this way, high level information for target localization and low level information for shape description can be fused and enhanced by our MFFM.} 

\subsubsection{Multi-level ViT Module}\label{MFEM}
MVTM contains ($k-1$) Vision Transformer (ViT) branches, and all branches have the same structure. we adopt the ResNet-18 \cite{38} as the feature embedding module to extract multi-scale local features $\bm{F_i}\in{{\mathbb{R}}^{{C_i}\times{H_i}\times{W_i}}}(i\in{\{1, 2, \dots, k\}})$. Features $\bm{F_i}$ are flattened into 2D patches $\bm{{E_{em}}^{(i)}}\in{{\mathbb{R}}^{{N_i}\times({{{P_i}^2}{C_i}})}}$, where $N_i = \frac{{H_i}{W_i}}{{P_i}^2}$ is the number of patches and $(P_i,P_i)$ is the resolution of each patch. After position embedding, we get $\bm{E^{(i)}_{pos}}$. We obtain embedded tokens $\bm{E^{(i)}} = \bm{E^{(i)}_{em}} +\bm{E^{(i)}_{pos}}$, where $n$ is the number of tokens, and $n = \frac{{H_i}{W_i}}{P_i^2}$. Embedded tokens $\bm{E^{(i)}}$ are divided into $m$ heads $\bm{E^{(i)}} =\{\bm{E^{(i)}_1}, \bm{E^{(i)}_2}, \dots, \bm{E^{(i)}_j}, \dots, \bm{E^{(i)}_m}\}$, $\bm{E^{(i)}_j}\in{\mathbb{R}^{n\times{\frac{C_i}{m}}}}$ ($j\in \{1, 2, \dots, m\}$), and then fed into the multi-head self-attention module $\rm{MSA}$ to obtain interaction tokens $\bm{E^{(i)}_a}$. We define these
processes as:
{\setlength\abovedisplayskip{0.25pt}
\setlength\belowdisplayskip{0.25pt}\begin{equation}\label{MSA}
{\bm{E^{(i)}_a} = {\rm{MSA}}[{\rm{LN}}(\bm{E^{(i)}})] + \bm{E^{(i)}}}, 
\end{equation}}
where $\rm{LN}$ is the layer normalization.

\par In each head, the multi-head self-attention module $\rm{MSA}$ defines three trainable weight matrices to transform Queries $\bm{Q^{(i)}}$, Keys $\bm{K^{(i)}}$ and Values $\bm{V^{(i)}}$. Then, $\bm{E^{(i)}_a}$ are fed into the $\rm{MLP}$ module to obtain final tokens $\bm{E^{(i)}_b}$, the result of $\rm{MLP}$ can be expressed as:
{\setlength\abovedisplayskip{0.25pt}
\setlength\belowdisplayskip{0.25pt}\begin{equation}\label{MLP}
{{\bm{E^{(i)}_b}} = {\rm{MLP}}[{\rm{LN}}({\bm{E^{(i)}_a}})] + {\bm{E^{(i)}_a}}}, 
\end{equation}}
where ${\bm{E^{(i)}_b}}\in{{\mathbb{R}}^{{N_i}\times({{{P_i}^2}{C_i}})}}$.

\par Then, we reshape tokens ${\bm{E^{(i)}_b}}\in{{\mathbb{R}}^{{{{P_i}^2}{C_i}}}}$ into features ${\bm{V}_i}$, where $P_i = H_k = W_k$.
The result of the ViT branch can be expressed as:
{\setlength\abovedisplayskip{0.1pt}
\setlength\belowdisplayskip{1pt}\begin{equation}\label{Vit_brance}
{{\bm{{V}_i}} = {\rm{Reshape}}\{{\rm{MLP}}[{\rm{LN}}({\bm{E^{(i)}_a}})]\} + {\bm{E^{(i)}_a}})}.
\end{equation}}
\begin{figure*}[t]
\centering
\setlength{\belowcaptionskip}{-0.3cm}
\includegraphics[width=18.2cm]{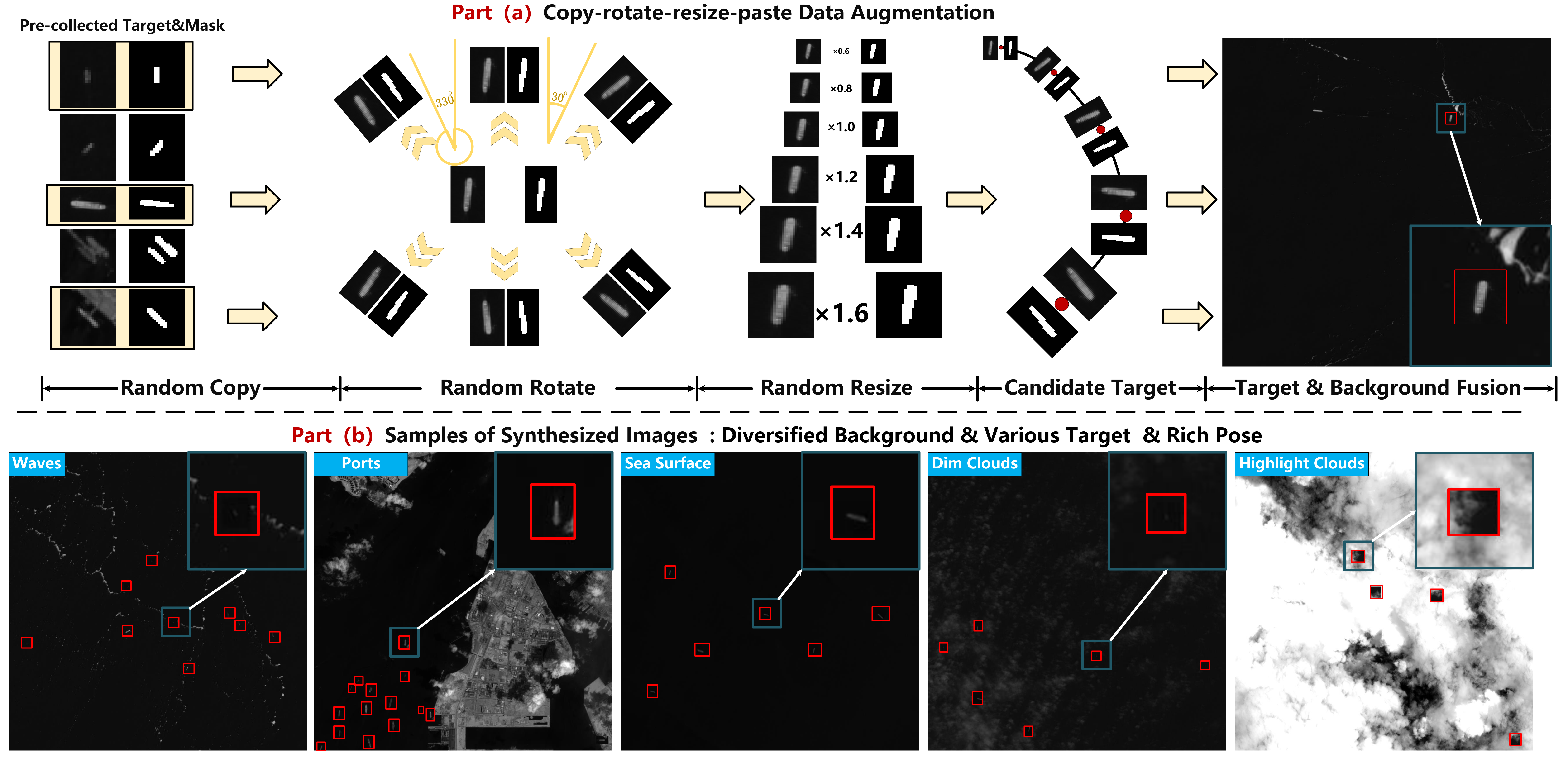}\caption{An illustration of the copy-rotate-resize-paste data augmentation method. (a) Copy-rotate-resize-paste data augmentation. Images of the targets' neighborhood are first collected and randomly copied. Then, the selected target is randomly rotated. After that, the target is randomly resized as a candidate target. Finally, the candidate target is pasted into the background area of image background region. (b) Samples of synthesized images.}\label{CRRP_overall}
\end{figure*}
\vspace{-1em}

\subsubsection{Multi-level Feature Fusion Module}\label{MFFM}
in our multi-level feature fusion module, features are fused by capturing long-distance dependency of these extracted high-level features. The multi-level features $\{\bm{{V}_i}\}(i\in{\{1, 2, \dots, k-1\}}$ and CNN local features $\bm{F_k}$ are first concatenated and then fed to {an} $1\times{1}$ convolution to fuse the feature $\bm{{M}_k}\in{{\mathbb{R}}^{{C_k}\times{H_k}\times{W_k}}}$. All multi-level features are fused by capturing long-distance dependency of these extracted high-level features. The fusion features can be expressed as:
{\setlength\abovedisplayskip{0.5pt}
\setlength\belowdisplayskip{1pt}\begin{equation}\label{Eq:MFFM}
{{\bm{M_k}} = {\rm{Conv}}[{\rm{Concat}}({\bm{{F}_k}}, {\bm{{V}_{k-1}}}, {\bm{{V}_{k-2}}}, \dots, {\bm{{V}_1}})]}.
\end{equation}}
\vspace{-2.1em}

\subsection{U-shape Decoder}\label{U-shape_decoder}
To obtain confidence maps of small targets, we adopt a decoder to upsample multi-level features $\bm{M_k}$. Multi-layer features $\{\bm{{F}_i}\}(i\in{\{1, 2, \dots, k-1\}}$ in the encoder are concatenated with features obtained by upsampling operation through skip connection operation to generate $\bm{M_i}(i\in{\{k-1,k-2,...,1\}})$. The processing of the decoder can be expressed as:
{\setlength\abovedisplayskip{0.25pt}
\setlength\belowdisplayskip{0.25pt}
\begin{equation}\label{Eq:U shape}
{{\bm{M_{i-1}}} = {\rm{Conv}}\{{\rm{Concat}}[{\bm{F_{i-1}}}, {\rm{Upsample}}({\bm{M_{i}}})]\}}.
\end{equation}}

\par A robust predicted probability map can be expressed as:
{\setlength\abovedisplayskip{0.5pt}
\setlength\belowdisplayskip{1pt}\begin{equation}\label{Eq:P}
{{\bm{P}} = {\rm{Sigmoid}}({\bm{M_0}})}.
\end{equation}}
\vspace{-2.2em}

\subsection{The Eight-connected Neighborhood Clustering Module}\label{Eight_Module}
After the U-shape decoder, we introduce an eight-connected neighborhood clustering module \cite{39} to clutter all pixels and calculate the centroid of each target. If any two pixels $(m_0, n_0)$, $(m_1, n_1)$ in feature maps $\bm{P}$ have intersection areas in their eight neighborhoods, i.e.,
{\setlength\abovedisplayskip{0.2pt}
\setlength\belowdisplayskip{0.2pt}
\begin{equation}\label{Eight1}
{{{\rm{N_{8}}}(m_0,n_0)} \cap {\rm{{N_{8}}}}(m_1,n_1) \neq 0},
\end{equation}}
where ${{\rm{N_{8}}}(m_0,n_0)}$ and ${\rm{{N_{8}}}}(m_1,n_1)$ represent the eight neighbor-hoods of pixel $(m_{0},n_{0})$ and $(m_{1},n_{1})$. Then, $(m_{0},n_{0})$ and $(m_{1},n_{1})$ are judged as adjacent pixels. If the these two pixels have the same value, i.e.,
{\begin{equation}\label{Eight2}
{{p(m_0,n_0)} = {p(m_1,n_1)}, \forall ({p(m_0,n_0)},{p(m_1,n_1)}) \in {\bm{P}}},
\end{equation}}
where ${p(m_0,n_0)}$ and $p(m_1,n_1)$ represent the value of pixel $(m_{0}, n_{0})$ and $(m_{1}, n_{1})$, these two pixels are considered to belong to the same target area. Once all targets in the image are determined, centroid can be calculated according to their coordinates.
\vspace{-1em}

\subsection{Data Augmentation}\label{Data Augmentation}
As mentioned in Section \ref{SecMotivations}, the distribution of the foreground targets and background are extremely imbalanced in our NUDT-SIRST-Sea. This foreground-background imbalance issue makes the network pay more attention to those uninformative background regions and thus hinders the quick convergence of the network. {Copy-paste (CP) data augmentation is a powerful data augmentation method for instance segmentation \cite{40}. Based on CP data augmentation method, we further propose a copy-rotate-resize-paste data augmentation method (namely, CRRP) to manually increase the ratio of candidate targets in the training phase and thus accelerate the convergence of the network. Our CRRP data augmentation method copies both targets and target neighborhood background while CP data augmentation method only copy targets. In this way, our CRRP data augmentation method can well preserve the information of target itself and contextual information between targets and background. Otherwise, suspicious targets (e.g. tiny clouds, port containers, reefs, and land bright spots) are detected as targets without the contextual dependency. Therefore, our CRRP is a more suitable data augmentation method for space-based SIRST detection task compared to the CP.}

\par As shown in Fig \ref{CRRP_overall} (a), we first collect images of the targets' neighborhood and randomly copy one target. Then, the selected targets are randomly rotated. After that, the target is randomly resized as a candidate target. Finally we paste the candidate target into the background area of image background region. As shown in Fig \ref{CRRP_overall} (b), the imbalance of the foreground targets and background distribution is relieved and the training time is also greatly reduced as compared to previous simple data augmentation methods (e.g., rotate, translate, color jitter).
\vspace{-2em}

\subsection{FocalIoU Loss}\label{FocalIoU loss}
Focal loss \cite{41} focuses on hard samples (e.g., small scale targets, edges of targets, and suspicious targets) which helps target localization. However, Focal loss causes more false alarm due to the high response in the background suspicious area. SoftIoU loss \cite{42} focuses on large-scale targets and lose small-scale targets. That is because, large-scale targets contributes much more in {intersection over union ($IoU$)} than small-scale targets, resulting in lost of small-scale targets. To achieve the `double-win' of target localization and shape description, we combine the SoftIoU loss and the Focal loss to develop a FocalIoU loss. Our FocalIoU loss combines the advantages of the Focal loss and the SoftIoU loss, with a low response in background areas and focus on small-scale targets. The formula of our FocalIoU loss function is expressed as Eq. \ref{FIoUloss}. 
\begin{equation}\label{Focal loss}
{{{\rm{FL}}(p,y)} = - y{(1 - p)^\gamma }{{\rm{log}}(p)} -{ (1 - y)}{p^\gamma }{{\rm{log}}(1 - p)}},
 \end{equation}
\begin{equation}\label{SoftIoU}
SoftIoU= \frac{smooth + \sum {p\times{y}} }{smooth + \sum {p + \sum {y - } } \sum {p\times{y} }},
\end{equation}
\begin{equation}\label{FIoUloss}
{{\rm{FIoUL}}(p,y) =2(1-SoftIoU){[{\rm{FL}}(p,y)]^\frac{1+SoftIoU}{2}}},
 \end{equation}
 where $p$ denotes the probability of each pixel, $y$ denotes the label of each pixel in probability map $\bm{P}$ and {$\gamma$ is an adjustable factor to control the attention on hard samples. $FL$ and $FIoUL$ are the abbreviations of Focal loss and FocalIoU loss respectively. $SoftIoU$ is a convergent $IoU$ with an adjustable factor $smooth$ to avoid infinity.} 
 
 \par To further analyze our FocalIoU loss function, we derive the FocalIoU loss function in Eq. \ref{Eq:dFocalIoU loss}
\begin{equation}\label{Eq:dFocal loss}
 \begin{aligned}
 \frac{{\partial}{{\rm{FL}}(p,y)}}{{\partial p}} = & -(1 - y)\gamma {p^{\gamma - 1}}{\rm{log}}(1 - p) + (1 - y){p^\gamma }\frac{1}{{1 - p}}\\
 &+y\gamma {(1 - p)^{\gamma - 1}}{\rm{log}}(p)-y{(1 - p)^\gamma }\frac{1}{p},
 \end{aligned}
 \end{equation}
 \begin{equation}\label{Eq:dFocalIoU loss}
 \begin{aligned}
 \frac{\partial{\rm{FIoUL}}(p,y)}{\partial x} =&{{(1-SoftIoU^2)}{{{\rm{FL}}(p,y)}^{\frac{1-SoftIoU}{2}}}}\\
 &\cdot {\frac{\partial \rm{FL}(p,y)}{\partial p}}\frac{\partial p}{\partial x},
 \end{aligned}
 \end{equation}
where $p = {\rm{Sigmoid}}(x)$, $x$ is the value of each pixel in the MTU-Net output map and $p$ is the probability value of each pixel.

\begin{figure}[t]
\centering
\setlength{\belowcaptionskip}{-0.5cm}
\includegraphics[width=8.8cm]{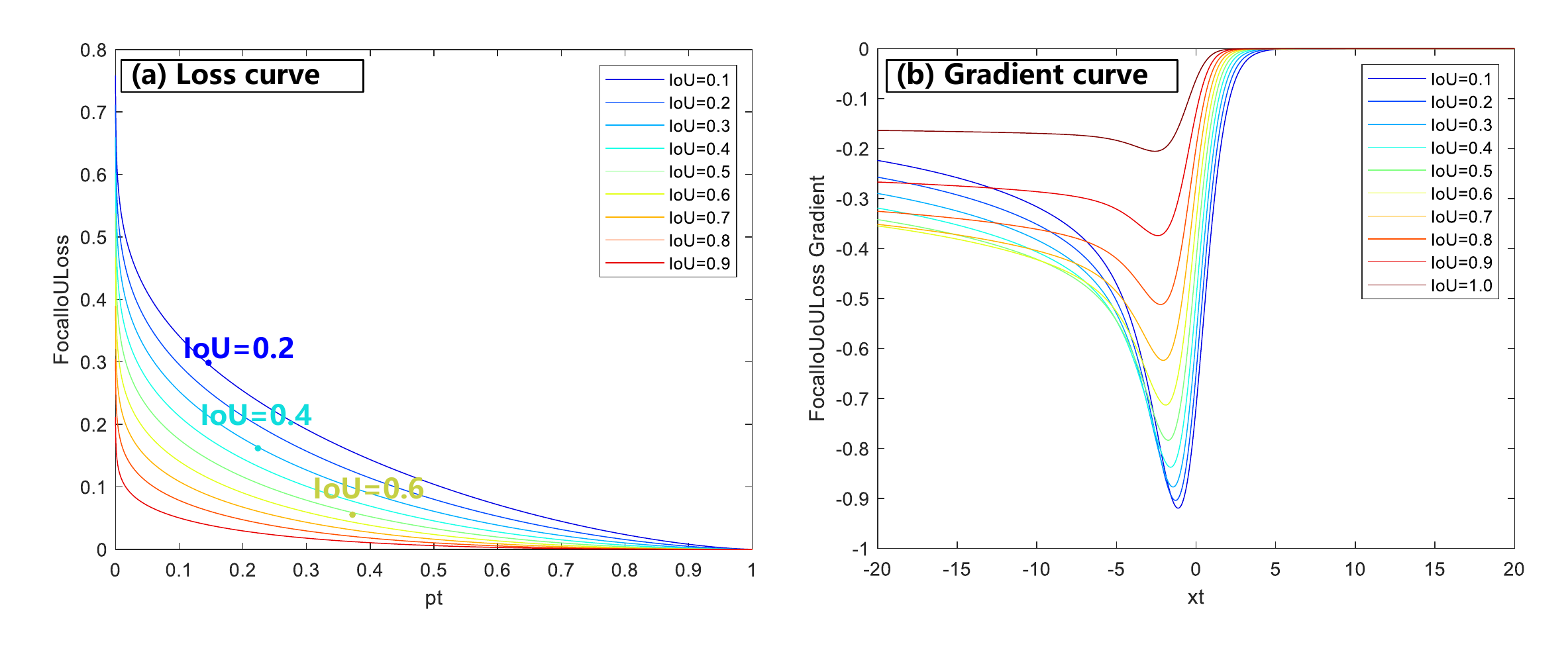}
\caption{FocalIoU loss analysis. (a) FocalIoU loss function curve. (b) FocalIoU loss function gradient curve.}\label{fig:FocalIoU loss_curve_}
\end{figure} 

\par As shown in Fig. \ref{fig:FocalIoU loss_curve_} (a), samples with low $IoU$ output result in a high FocalIoU loss and a sharp decrease of the FocalIoU loss. When $IoU$ is small, the overall segmentation performance of this image is poor, and the FocalIoU loss {focuses on difficult simple samples} (e.g., large-scale targets) more than difficult samples. Consequently, $F_a$ decreases while $IoU$ increases. When $IoU$ is large, the FocalIoU loss performs like the Focal loss and focuses more on difficult samples, which helps $P_d$ to increase.

\section{Experiment}\label{SecExperiment}
In this section, we first introduce our evaluation metrics and implementation details. Then, we compare our MTU-Net to several state-of-the-art SIRST detection methods. Finally, we present ablation studies to investigate our network.
\vspace{-1em}
\subsection{Evaluation Metrics}\label{Evaluation Metrics}
Following the pioneering DNA-Net \cite{27}, we adopt probability of detection (${P}_{d}$) and false alarm rate (${F}_{a}$) to evaluate the localization performance and use intersection over union ($IoU$) to evaluate shape description performance. {Besides, we adopt receiver operating characteristic \cite{43} (ROC) analysis to further show the overall detection effectiveness, target detect ability, and background suppression ability}
\subsubsection{Probability of Detection}
Probability of detection (${P}_{d}$) is a target-level evaluation metric. It measures the ratio of correctly predicted target number $T_{correct}$ over all target number $T_{All}$. ${P}_{d}$ is defined as:
{\begin{equation}\label{PD}
 {P}_{d}= \frac{T_{correct}}{T_{All}}.
 \end{equation}}
 
\par If the centroid deviation of the target is smaller than the pre-defined deviation threshold $D_{\textit{thresh}}$, we consider those targets as correctly predicted ones. We set the pre-defined deviation threshold as $3$ in this paper.

\subsubsection{False Alarm Rate}
False alarm rate (${F}_{a}$) is another target-level evaluation metric. It is used to measure the ratio of falsely predicted pixels $P_{false}$ over all image pixels $P_{All}$. ${F}_{a}$ is defined as:
 {\begin{equation}\label{FA}
 {F}_{a} = \frac{P_{false}}{P_{All}}.
 \end{equation}}
\par If the centroid deviation of the target is larger than the pre-defined deviation threshold, we consider those pixels as falsely predicted ones.

\subsubsection{Intersection over Union}
Intersection over Union ($IoU$) is a target-level evaluation metric. It evaluates the target description performance of an algorithm. $IoU$ is calculated as the ratio of intersection and the union areas between the target-predictions and target-labels. $IoU$ is defined as:
 {\begin{equation}\label{IoU}
 {{IoU}= \frac{Target_{inter}}{Target_{Union}},}
 \end{equation}where $Target_{inter}$ and $Target_{Union}$ represent the interaction areas and {union areas} between target-prediction and target-label, respectively.}

\subsubsection{{3D ROC}}
{
 Receiver operating characteristic \cite{43} (ROC) analysis is widely applied in object detection field. Common ROC curves generally include a 3D ROC curve specified by threshold $\tau$, probability of detection $P_d$ and false alarm rate $F_a$, and three 2D ROC curves of ($\tau$,$P_d$), ($\tau$,$P_d(\tau)$), and ($\tau$,$F_a(\tau)$). Both $P_d(\tau)$ and $F_a(\tau)$ can be calculated in Eq. \ref{PD} and Eq. \ref{FA} by changing predicted pixel threshold $\tau$. The 2D ROC curves of ($F_a$,$P_d$), ($\tau$,$P_d$), and ($\tau$,$F_a$) indicate overall detection effectiveness, target detect ability, and background suppression ability of different methods, respectively. A good detector has 2D ROC curves of ($F_a$,$P_d$), ($\tau$,$P_d$), and ($\tau$,$F_a$) close to the upper left, upper right, and lower left corner of the coordinate axis.}
\subsection{Implementation Details}\label{Protocol}
{NUDT-SIRST-Sea dataset contains $41$ images for training and $7$ images for testing. These real images were captured by sensors mounted on a low earth-orbiting satellite. 
All input images with resolution of $10000\times10000$ were first cut into patches with a resolution of $1024\times1024$.} {Before training, all input images were first normalized. Then, these normalized images were sequentially processed by random image flip, Gaussian blurring, and CRRP for data augmentation before being fed into the network.} ResNet-18\cite{38} was chosen as our segmentation backbone. The number of down-sampling layers $i$ was $4$. {Our network was trained using the FocalIoU loss function and optimized by the Adagrad method \cite{44} with the CosineAnnealingLR scheduler. We initialized the weights and bias of our model using the Xavier method \cite{45}. We set the learning rate, batch size, and epoch as $0.05$, $8$, and $1500$, respectively.} All models were implemented in PyTorch \cite{46} on a computer with an AMD Ryzen 9 3950X @ $2.20$ GHz CPU and {a} Nvidia RTX 3090 GPU.
\begin{figure*}[t]
 \centering
 \setlength{\belowcaptionskip}{0.2cm}
 \includegraphics[width=18.2cm]{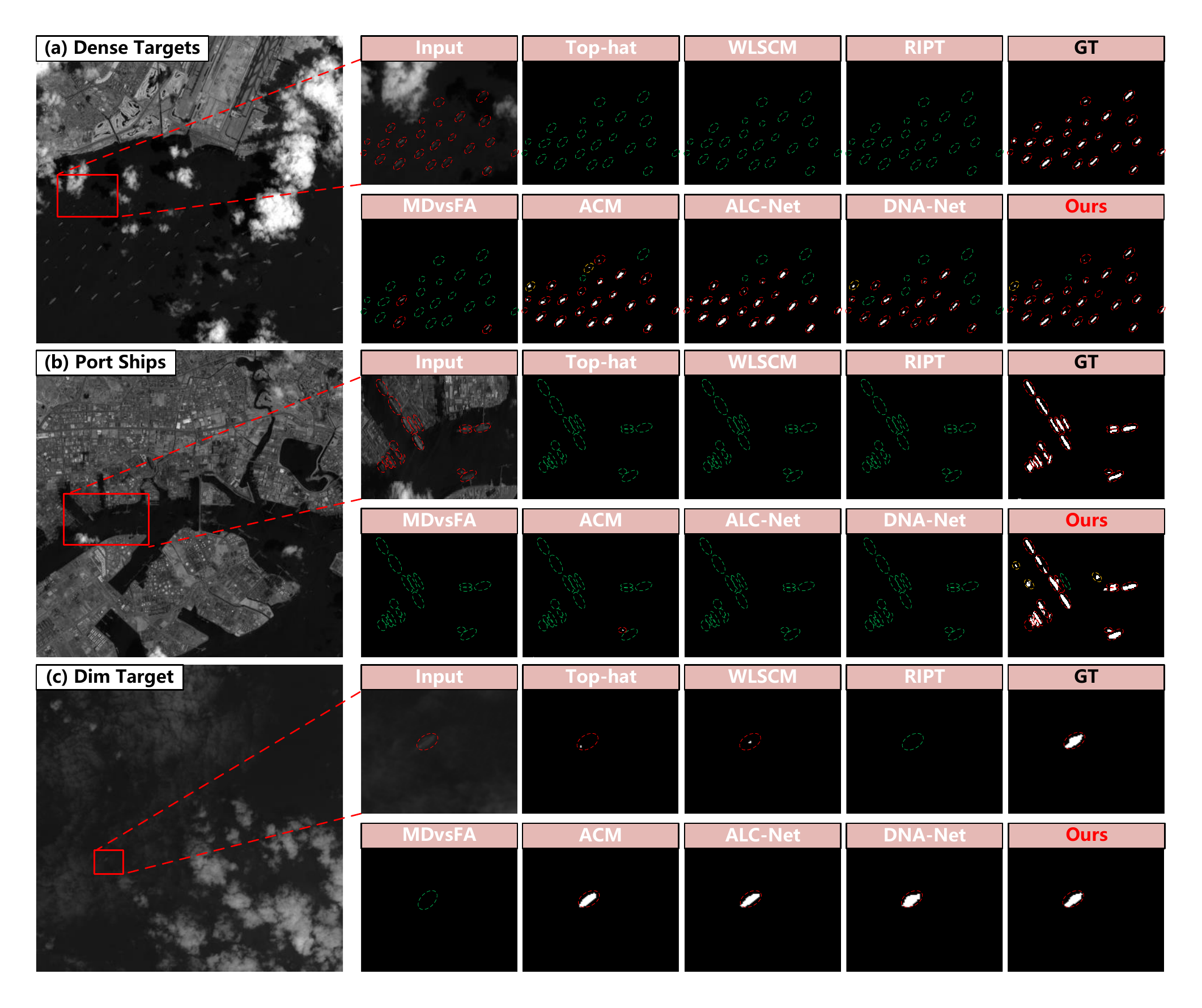}
 \caption{Qualitative results achieved by different SIRST detection methods in three typical scenes (e.g., dense targets, port ships, dim targets). For better visualization, the target area is enlarged. The correctly detected target, false alarm, and miss detection areas are highlighted by \textcolor{red}{red}, \textcolor{yellow!40!orange}{yellow}, and \textcolor{green}{green} dotted circles, respectively. Our MTU-Net can generate output with precise target localization and shape segmentation under a smaller $F_a$.}\label{fig:SOTA}
\end{figure*}
\begin{table*}[t]
\setlength{\belowcaptionskip}{2cm}
\label{SOTA_table}
\centering
\renewcommand\arraystretch{1.2}
\caption{$IoU$, $P_{d}$, and $F_{a}$ values achieved by different traditional and CNN-based state-of-the-art methods on the NUDT-SIRST-Sea dataset. For $IoU$ and $P_{d}$, larger values indicate higher performance. For $F_{a}$, smaller values indicate higher performance. The best results are in \textcolor{red} {red} and the second best results are in  {blue}.} \label{TabOVER_ALL}
\begin{tabular}{|l|c|c|c|l|c|c|c|c}
\hline
\multicolumn{4}{|c|}{Traditional methods} & \multicolumn{4}{c|}{Deep learning based methods} \\ \hline
\multicolumn{1}{|c|}{\multirow{2}{*}{Method Description}} & {$IoU$}& {$P_{d}$}&{$F_{a}$} &\multicolumn{1}{|c|}{\multirow{2}{*}{Method Description}} &{$IoU$}& {$P_{d}$}&{$F_{a}$} \\&{($\times10^{-2}$)} &{($\times10^{-2}$)}&{($\times10^{-6}$)}&{ }&{($\times10^{-2}$)} &{($\times10^{-2}$)}&{($\times10^{-6}$)} \\ \hline
Filtering Based: Top-Hat\cite{6} & $1.17$ & $2.68$ & $86.03$ 
&CNN Based: ACM\cite{23} &$47.57$ & $70.46$ & $21.31$ \\ \hline
Filtering Based: Max-Median\cite{7} & $0.28$ &$1.04$ & $87.47$ 
&CNN Based: ALC-Net\cite{24}& $48.9$ & $58.65$ & \textcolor{blue}{$\mathbf{9.13}$} \\ \hline
Local Contrast Based: WSLCM\cite{11} & $0.60$ & $10.52$ & $87.25$
&CNN Based: MDvsFA-cGAN\cite{25}& $0.37$ & $0.18$ & $92$ \\ \hline
Local Contrast Based: TLLCM\cite{10} &$0.59$ & $5.66$ & $87.22$ 
&CNN Based: ResU-Net\cite{47} & $41.05$ & $60.18$ &\textcolor{red}{$\mathbf{7.92}$} \\ \hline
Local Rank Based: MSLSTIPT\cite{5} & $0.33$ & $0.35$ & $83.55$ 
&CNN Based: DNA-Net\cite{27} & $42.17$ & $61.60$ &$17.19$ \\ \hline
Local Rank Based: NRAM\cite{15} &$0.35$ & $17.31$ & $87.49$ 
&\textbf{MTU-Net-ResNet10 (ours)} & $59.98$ & $81.22$ & $16.64$ \\ \hline
Local Rank Based: RIPT\cite{16} & $0.36$ &$26.08$ &$87.48$ 
&\textbf{MTU-Net-ResNet18 (ours)} & \textcolor{red}{$\mathbf{64.14}$} & \textcolor{red}{$\mathbf{85.44}$} & {$11.72$} \\ \hline
Local Rank Based: PSTNN\cite{17} & $1.50$ &$13.51$ &$86.34$
&\textbf{MTU-Net-ResNet34 (ours)} & \textcolor{blue}{$\mathbf{62.00}$} & \textcolor{blue}{$\mathbf{84.70}$} & {$12.12$} \\ \hline
\end{tabular}
\end{table*}
\subsection{Comparison to the State-of-the-art Methods}\label{Comparison}
 To demonstrate the superiority of our method, we compare our MTU-Net with several state-of-the-art (SOTA) methods, including traditional methods (filtering based methods: Top-Hat \cite{6} and Max-Median \cite{7}); local contrast based methods: TLLCM \cite{10} and WSLCM \cite{11}); local rank based methods: NRAM \cite{15}, RIPT \cite{16} and PSTNN \cite{17}) and CNN-based methods including DNA-Net\cite{27}, MDvsFA-cGAN \cite{25}, ACM \cite{23}, ALC-Net \cite{24} and ResU-Net \cite{47}) on the NUDT-SIRST-Sea dataset. For fair comparison, we retrained all the CNN-based methods on our NUDT-SIRST-Sea dataset.
\subsubsection{Qualitative Results}
Qualitative results on our NUDT-SIRST-Sea are shown in Fig. \ref{fig:SOTA}. Compared with traditional methods, our method can generate more precise localization and classification results with smaller $F_a$. The results achieved by traditional methods easily lose dense small-scale targets (Fig. \ref{fig:SOTA} (a)) and targets in port (Fig. \ref{fig:SOTA} (b)). Traditional methods generate bad shape segmentation in dim targets (Fig. \ref{fig:SOTA} (c)). The CNN-based methods (MDvsFA-cGAN, ResU-Net, ACM, ALC-Net, DNA-Net) perform much better than traditional methods. However, due to the extreme dim targets (Fig. \ref{fig:SOTA} (a), Fig. \ref{fig:SOTA} (c)) in our NUDT-SIRST-Sea, MDvsFA-cGAN loses more targets. Our MTU-Net can generate better shape segmentation (Fig. \ref{fig:SOTA} (c)) than DNA-Net, ACM and ALC-Net. Our MTU-Net can generate better target localization in scenes with port ships (Fig. \ref{fig:SOTA} (b)). That is because, our MTU-Net can effectively capture long-distance dependency with the help of coarse-to-fine multi-level ViT module.
\subsubsection{Quantitative Results}
Similar to DNA-Net \cite{27}, we first obtained their predicts and then performed noise suppression by setting a threshold to remove low-response areas for all the compared algorithms. Specifically, the adaptive threshold (${T}_{adaptive}$) was calculated for traditional methods according to:
\begin{equation}\label{threshold}
{\emph{T}_{adaptive}={\rm{Max}}[0.7{\rm{Max}}(\bm{P}), 0.5{\rm{\sigma}}(\bm{P})+{\rm{Avg}}(\bm{P})]},
\end{equation}
where ${\rm{Max}}(\bm{P})$ represents the largest value of output, ${T}_{adaptive}$ is an adaptive threshold, ${\rm{\sigma}}(\bm{P})$ and ${\rm{avg}}(\bm{P})$ denote the standard derivation and average value of output, respectively.

\par For deep learning based methods, we followed their original papers and adopted their fixed thresholds (i.e., $0$, $0$, $0$, $0$, $0.5$ for DNA-Net\cite{27}, ResU-Net\cite{47}, ACM\cite{23}, ALC-Net\cite{24}, and MDvsFA-cGAN\cite{25}, respectively). We kept all remaining parameters the same as their original papers.

\par Quantitative results are shown in Table \ref{TabOVER_ALL}. Our MTU-Net outperforms traditional methods significantly. That is because, NUDT-SIRST-Sea contains challenging images with various scales, orientations and brightness of tiny ship targets. Our MTU-Net can effectively capture long-distance features between background and targets. Limited by manually-selected parameters, those model-driven traditional methods cannot well cope with such challenging scenes. It is worth noting that the improvements achieved by MTU-Net over other deep learning based methods (i.e., MDvsFA-cGAN, ACM, ALC-Net and DNA-Net) are obvious. Our MTU-Net achieves $64.14\%$ on $IoU$, $85.44\%$ on $P_d$ and $11.72\times10^{-6}$ on $F_a$. Our MTU-Net outperforms other deep learning methods more than $15\%$ on $IoU$ and $P_d$. Besides, our MTU-Net only suffers a decrease of $3.82\times10^{-6}$ in term of $F_a$ than ResU-Net. 
That is because, our MTU-Net can effectively capture long-distance dependency with the help of coarse-to-fine multi-level ViT module. The long-distance dependency helps network detect small targets and generate less false alarm in suspicious targets.
\subsubsection{{3D ROC Analysis}}
\begin{figure}[t]
\centering
 \setlength{\belowcaptionskip}{-0.5cm}
\includegraphics[width=8.8cm]{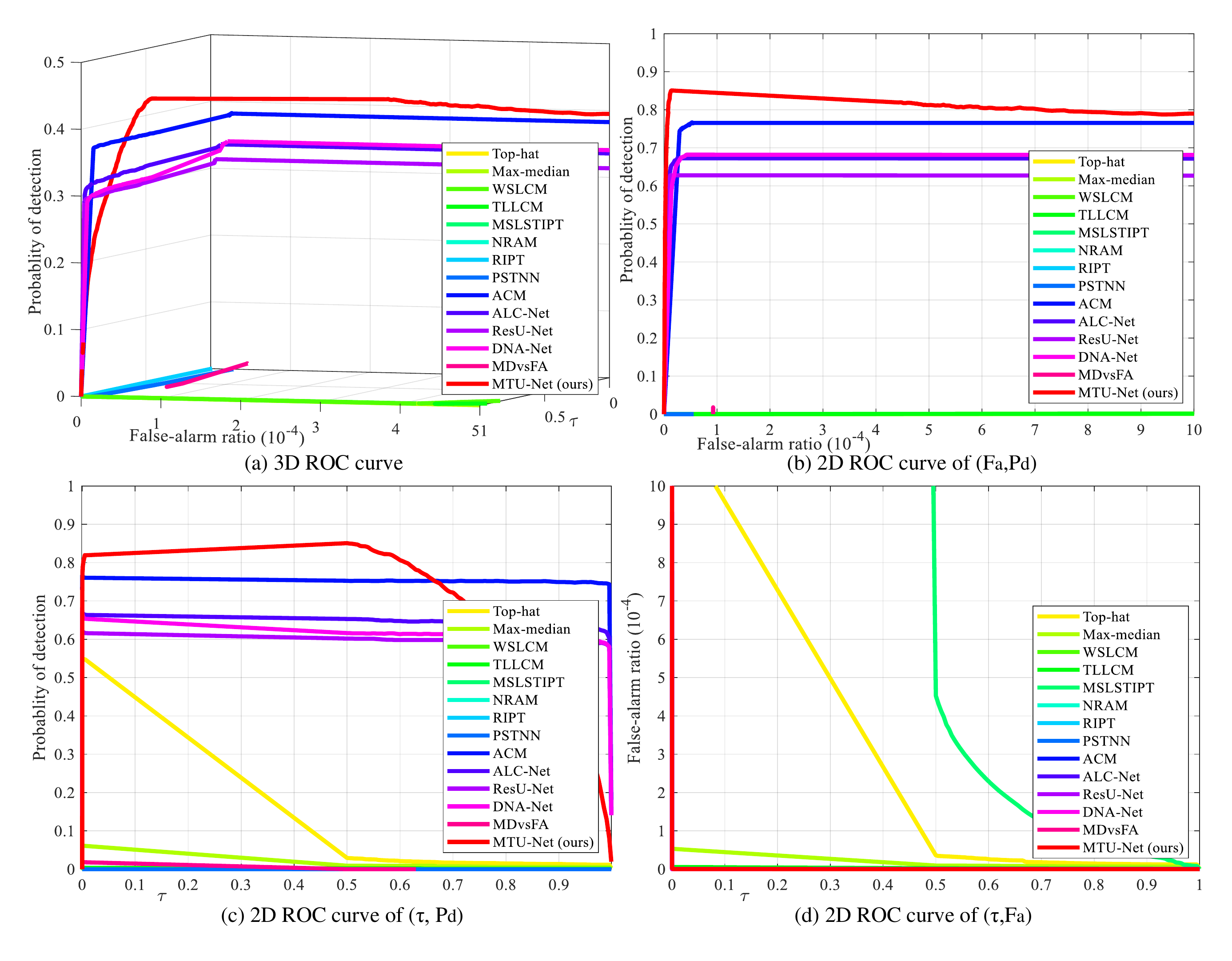}
\caption{{3D ROC and three corresponding 2D ROC curves of different comparing methods. (a) 3D ROC curve. (b) 2D ROC curve of ($F_a$,$P_d$). The curve of ($F_a$,$P_d$) closer to the upper left corner of the coordinate axis indicates a higher detection effectiveness. (c) 2D ROC curve of ($\tau$,$P_d$). The curve of ($\tau$,$P_d$) closer to the upper right corner of the coordinate axis indicates a better target detect ability. (d) 2D ROC curve of ($\tau$,$F_a$). The curve of ($\tau$,$F_a$) closer to the lower left corner of the coordinate axis indicates a better background suppression ability.}
}\label{fig:ROC}
\end{figure} 
{
Fig. \ref{fig:ROC} shows four kinds of ROC curves corresponding to
the detection maps. It can be observed that 2D ROC curve
of ($F_a$,$P_d$) of our MTU-Net is much closer to the upper left
corner than that of other methods in Fig. \ref{fig:ROC} (b). 2D ROC curves
of ($F_a$,$P_d$) show that our MTU-Net has a better detection
effectiveness than other methods. The 2D ROC curve of ($\tau$,$F_a$) of our MTU-Net in Fig. \ref{fig:ROC} (c) achieves a much higher $P_d$ when $\tau$ is smaller than $0.6$. 2D ROC curves
of ($\tau$,$F_a$) in Fig. \ref{fig:ROC} (b) show that our MTU-Net has an much better detection background suppression ability than traditional methods and the same the ability as other deep learning based methods. Our MTU-Net achieves an overall best performance on detection effectiveness, target detect ability, and background suppression ability.}
\subsection{Ablation Study}\label{Ablation Study}
In this subsection, we compare our MTU-Net with several variants to investigate the potential benefits introduced by our multi-level ViT module (MVTM), CRRP data augmentation and FocalIoU loss function. The results are shown in Table \ref{tab:branchs}, \ref{tab:CRRP}, \ref{loss_results}.
\vspace{0.5em}
\subsubsection{Multi-level ViT Module}\label{MRFEM}
the multi-level ViT module is used to achieve coarse-to-fine feature extraction. In our MVTM, ViT refines CNN features by capturing long-distance dependency of these extracted high level features. To demonstrate the effectiveness of our MVTM, we introduced the following network variants:
\begin{itemize}
\item \textbf{MTU-Net w/o level $\bm k$ ViT:} level $k$ denotes that we removed $1\sim k$ ViT branches from our MVTM. For fair comparison, we made model size comparable and retrained these variants in NUDT-SIRST-Sea;
\item \textbf{MTU-Net w/o MVTM:} we removed MVTM from our MTU-Net. For fair comparison, we made model size comparable and retrained this variant on NUDT-SIRST-Sea.
\end{itemize}
\vspace{0.5em}
\begin{table}[t]
\setlength{\belowcaptionskip}{0.5cm}
\caption{$IoU$, $P_{d}$, and $F_{a}$ values achieved by main variants
of MTU-Net on the NUDT-SIRST-Sea dataset.}
\label{table}
\small
\setlength{\tabcolsep}{3pt}
\centering
\begin{tabular}{|l|c|c|c|}
\hline
\multicolumn{1}{|c|}{\multirow{2}{*}{Model}}& 
{$IoU$}&{$P_{d}$}&{$F_{a}$}\\&{($\times10^{-2}$)} & {($\times10^{-2}$)} & {($\times10^{-6}$)} \\\hline
MTU-Net & 
{$\mathbf{64.14}$}& 
{$\mathbf{85.44}$}& 
{$\mathbf{11.72}$} \\
\hline
MTU-Net w/o level $1$ ViT& 
{${60.21}$}& {${82.59}$} &{${12.58}$} \\\hline
MTU-Net w/o level $2$ ViT& 
$59.68$& $82.85$&$15.86$ \\ \hline
MTU-Net w/o level $3$ ViT&
$53.29$& 
$81.96$ &$28.93$ \\\hline
MTU-Net w/o MVTM& 
$52.94$& 
$78.38$&$43.76$ \\\hline
\end{tabular}
\label{tab:branchs}
\end{table}

\begin{table}[t]
\caption{$IoU$, $P_{d}$, and $F_{a}$ values achieved by main variants of data augmentation used by MTU-Net on the NUDT-SIRST-Sea dataset. $\rm{CP}$ representatives using the copy-paste method for data augmentation.}
\label{tab:CRRP}
\small
\setlength{\tabcolsep}{3pt}
\centering
\begin{tabular}{|l|c|c|c|}
\hline
\multicolumn{1}{|c|}{\multirow{2}{*}{Model}}& 
{$IoU$}&{$P_{d}$}&{$F_{a}$}\\&{($\times10^{-2}$)} & {($\times10^{-2}$)} & {($\times10^{-6}$)} \\
\hline
MTU-Net w/o DA& 
$58.97$& 
$80.39$&$23.68$ \\\hline
MTU-Net with CP DA& 
{${61.45}$}&{${82.64}$}
&{${17.24}$}\\\hline
MTU-Net with CRRP DA& 
{$\mathbf{64.14}$}& 
{$\mathbf{85.44}$}& 
{$\mathbf{11.72}$} \\\hline
\end{tabular}
\end{table}

\par As shown in Table \ref{tab:branchs}, MTU-Net w/o level $1$ ViT suffers decreases
of $3.93\%$, $2.85\%$, and an increase of $0.86\times10^{-6}$ in terms of $IoU$, $P_d$, and $F_a$ values over MTU-Net on the NUDT-SIRST-Sea dataset. As the number of ViT branch decreases, the values of MTU-Net in $IoU$ and $P_d$ gradually decrease and the value of $F_a$ gradually increases. That is because, fewer multi-level features are extracted by MVTM in the multi-level ViT CNN hybrid encoder. Since fewer long distance information is used, the performance is poor. Specifically, when all {ViT branches} are pruned and MVTM is removed, MTU-Net suffers decreases of $11.20\%$, $7.06\%$, and an increase of $32.04\times10^{-6}$ in terms of $IoU$, $P_d$, and $F_a$ values.
\begin{figure*}[t]
 \centering
 \setlength{\belowcaptionskip}{-0.5cm}
 \includegraphics[width=18.2cm]{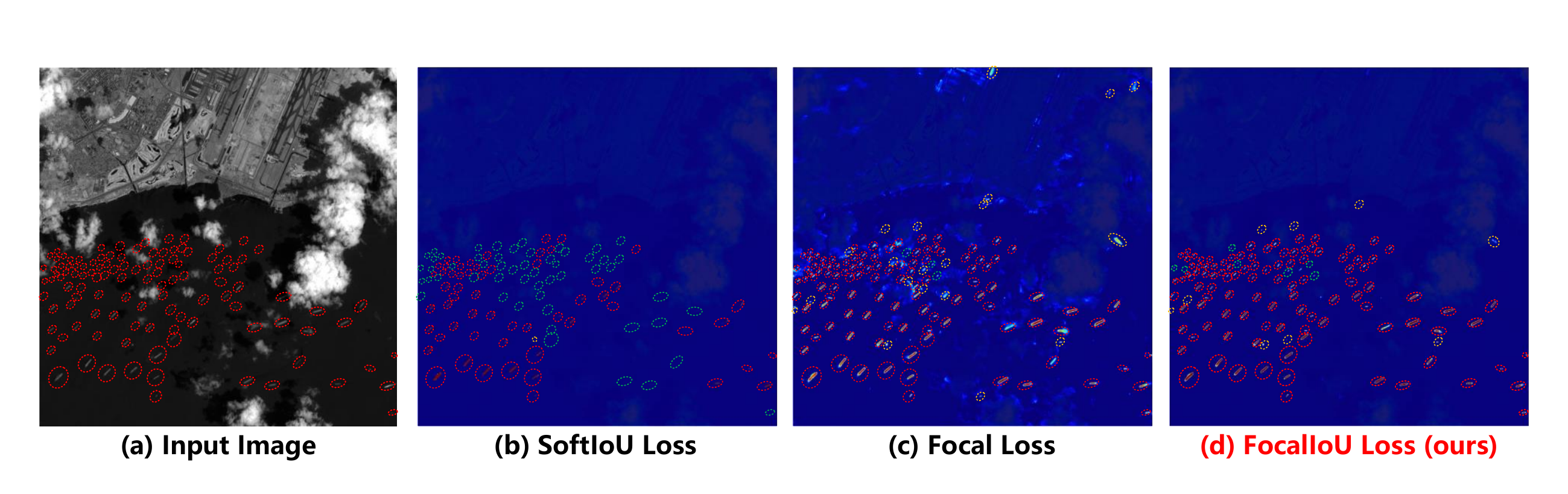}
 \caption{Visualization maps of MTU-Net output. The output of MTU-Net is marked by a solid red frame. The correctly detected target, false alarm, and miss detection areas are highlighted by \textcolor{red}{red}, \textcolor{yellow!40!orange}{yellow}, and \textcolor{green}{green} dotted circles, respectively. (a) The Input Image. (b) The final probability density map from the output of MTU-Net (SoftIoU loss). The map shows the responses low in background area and loses more small scale targets. (c) The final probability density map from the output of MTU-Net (Focal loss). The map
 shows the output of MTU-Net (Focal loss) responses more in small scale targets area and causes more false alarm. (d) The final probability density map of MTU-Net (FocalIoU loss). The map has a low response in the background area and focus on small-scale targets. }\label{fig:FocalIoU loss_visualization}
 \end{figure*}
 \vspace{0.5em}
\subsubsection{Copy-Rotate-Resize-Paste Data Augmentation}\label{CRRP DA}
Since the distribution of the labels and background of the dataset is extremely imbalanced. This problem misleads the network to focus more on the background region of the image. The imbalance causes more false alarm and reduces the convergence speed. Therefore, we use a data augmentation method as a parameter-free solution to alleviate this problem.
\begin{itemize}
\item \textbf{MTU-Net w/o DA:} we removed CRRP data augmentation in this variant and retrained our MTU-Net on NUDT-SIRST-Sea;
\item \textbf{MTU-Net with CP-DA:} we used the copy-paste data augmentation method in this variant and retrained our MTU-Net on NUDT-SIRST-Sea;
\item \textbf{MTU-Net with CRRP-DA:} we used the copy-rotate-resize-paste data augmentation method in this variant and retrained our MTU-Net on NUDT-SIRST-Sea.
\end{itemize}
\par As shown in Table \ref{tab:CRRP}, MTU-Net with CP-DA suffers decreases of $2.69\%$, $2.80\%$, and an increase of $5.52\times10^{-6}$ in terms of $IoU$, $P_d$, and $F_a$ values over MTU-Net with CRRP-DA on our NUDT-SIRST-Sea dataset. MTU-Net w/o DA suffers decreases of $5.17\%$, $5.05\%$, and an increase of $11.96\times10^{-6}$ in terms of $IoU$, $P_d$, and $F_a$ values over MTU-Net with CRRP-DA on our NUDT-SIRST-Sea dataset. {Note that, the $F_a$ of the MTU-Net drops a lot using our CRRP data augmentation method. That is because, there are a large number of highlighted complex backgrounds and suspicious targets. These highlighted complex backgrounds and suspicious targets occupy much more area than real targets in space-based infrared image. Without data augmentation, MTU-Net causes more false alarms on highlighted complex backgrounds and suspicious targets. Our CRRP data augmentation method can preserve the long-range information and contextual information of targets. Thus, MTU-Net can better learn the long-range information of targets and achieve a better performance on $IoU$, $P_d$ and $F_a$. }

\subsubsection{FocalIoU Loss}
The FocalIoU loss helps MTU-Net focus more on images with low $IoU$, and reduces the weights of difficult samples relative to simple samples when $IoU$ is small. FocalIoU loss achieves the `double-win' of target localization and shape description. To demonstrate the effectiveness of our FocalIoU loss, we {retrained} our MTU-Net using the SoftIoU loss and the Focal loss for fair comparison.

\par Visualization maps shown in Fig. \ref{fig:FocalIoU loss_visualization} also demonstrate the effectiveness of our FocalIoU loss. The Focal loss focuses on hard samples (e.g., small-scale targets, the edges of targets). However, the Focal loss causes a higher response in the background area, resulting in more false alarm. The SoftIoU loss focuses on large-scale targets and loses small-scale targets because large-scale targets contribute much more in $IoU$ than small-scale targets, resulting in miss detection of small-scale targets. The FocalIoU loss combines the advantages of both Focal loss and SoftIoU loss, with a low response in the background area and focuses on small-scale targets.
\vspace{0.5em}
\begin{table}[t]
\caption{$IoU$, $P_{d}$, and $F_{a}$ values achieved by different loss functions used with MTU-Net on the NUDT-SIRST-Sea dataset.}
\label{loss_results}
\small
\setlength{\tabcolsep}{3pt}
\centering
\begin{tabular}{|p{75pt}|p{50pt}<{\centering}|p{45pt}<{\centering}|p{45pt}<{\centering}|}
\hline
\multicolumn{1}{|c|}{Loss Function}& 
{$IoU$($\times10^{-2}$)} & {$P_{d}$($\times10^{-2}$)} & {$F_{a}$($\times10^{-6}$)} \\ \hline
Focal loss\cite{41} & 
$53.16$& 
{$\mathbf{86.18}$}& 
$33.93$ \\\hline
SoftIoU loss\cite{42}& 
{${62.00}$}& 
$75.22$& 
{$\mathbf{9.42}$} \\\hline
FocalIoU loss (ours) & 
{$\mathbf{64.14}$}& 
{${85.44}$}& 
{${11.72}$} \\\hline
\end{tabular}
\end{table}
\vspace{0.5em}
\begin{figure}[t]
\centering
 \setlength{\belowcaptionskip}{-1cm}
\includegraphics[width=8.8cm]{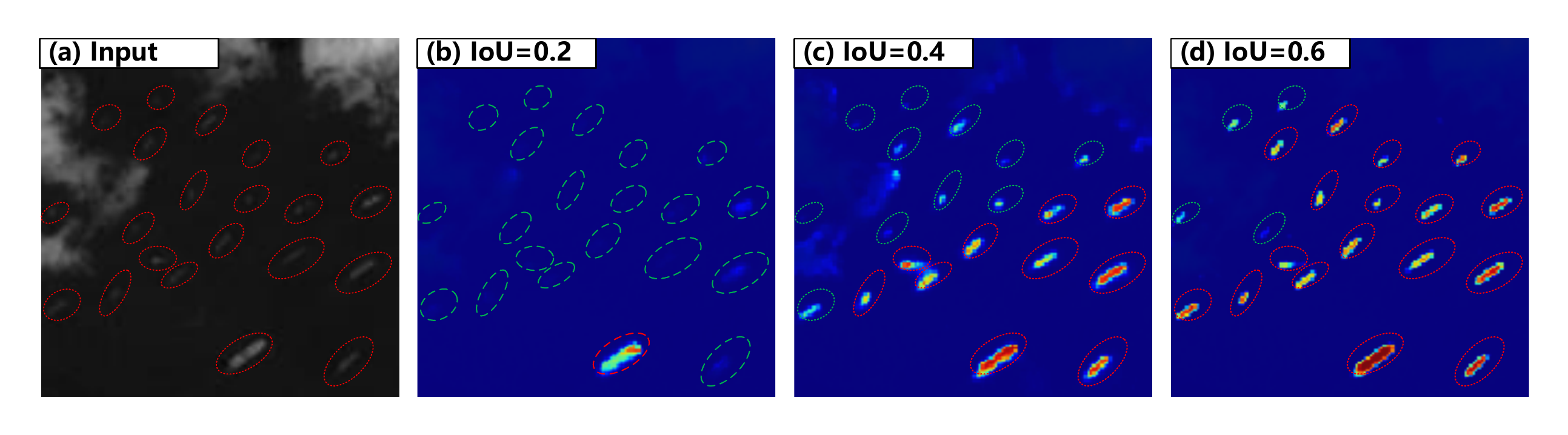}
\caption{FocalIoU loss analysis. (a) Input image. (b) (c) (d) Visualization maps of MTU-Net output in different $IoU$. The output of MTU-Net is marked by a solid red frame. The correctly detected target, false alarm, and miss detection areas are highlighted by \textcolor{red}{red}, \textcolor{yellow!40!orange}{yellow}, and \textcolor{green}{green} dotted circles, respectively. MTU-Net shifts from focusing on large scale targets to focusing on small scale targets as $IoU$ rises from $0.2$ to $0.6$. }\label{fig:FocalIoU loss_curve}
\end{figure} 
\par As shown in Table \ref{loss_results}, MTU-Net with the Focal loss suffers a decrease of $11.02\%$ and an increase of $22.21\times10^{-6}$ in terms of $IoU$ and $F_a$ values over MTU-Net with the FocalIoU loss. MTU-Net with the Focal loss achieves an increase of $0.74\%$ in $P_d$ value. That is because, the Focal loss focuses on difficult positive samples (e.g., small scale targets) but leads to high $F_a$ value and low $IoU$ value. MTU-Net with the SoftIoU loss suffers decreases of $2.14\%$ and $10.22\%$ in terms of $IoU$ and $P_d$ values over MTU-Net with the FocalIoU loss. MTU-Net with the SoftIoU loss achieves a decrease of $2.3\times10^{-6}$ in $F_a$ value. That is because, SoftIoU loss is calculated by the $IoU$ of output, leading to more focus on large scale targets. Numerous small-scale targets contribute less to $IoU$, resulting in higher $IoU$, smaller $F_a$ but smaller $P_d$.
\par As shown in Fig. \ref{fig:FocalIoU loss_curve}, MTU-Net shifts from focusing on large-scale targets to focusing on small-scale targets as $IoU$ rises. The above results demonstrate that our FocalIoU loss can achieve the `double-win' of target localization and shape description.

\section{CONCLUSION}\label{SecConclusion}
In this paper, we propose the first and largest manually annotated dataset for space-based infrared tiny ship detection. Besides, we propose a novel pipeline for space-based infrared tiny ship detection, which contains MTU-Net, the CRRP data augmentation method, and FocalIoU loss. Specifically, a multi-level feature extraction module is designed to adaptively extract multi-level long-distance features in our MTU-Net. The CRRP data augmentation method is designed to alleviate the imbalance between target and background samples. FocalIoU loss is proposed to achieve accurate target localization and shape description.
Experimental results on the NUDT-SIRST-Sea dataset show that the proposed MTU-Net model outperforms traditional SIRST methods and existing deep learning based SIRST methods in a set of evaluation metrics.
\bibliographystyle{IEEEtran}
\bibliography{MTU-Net}

\begin{IEEEbiography}[{\includegraphics[width=1in,height=1.25in,clip,keepaspectratio]{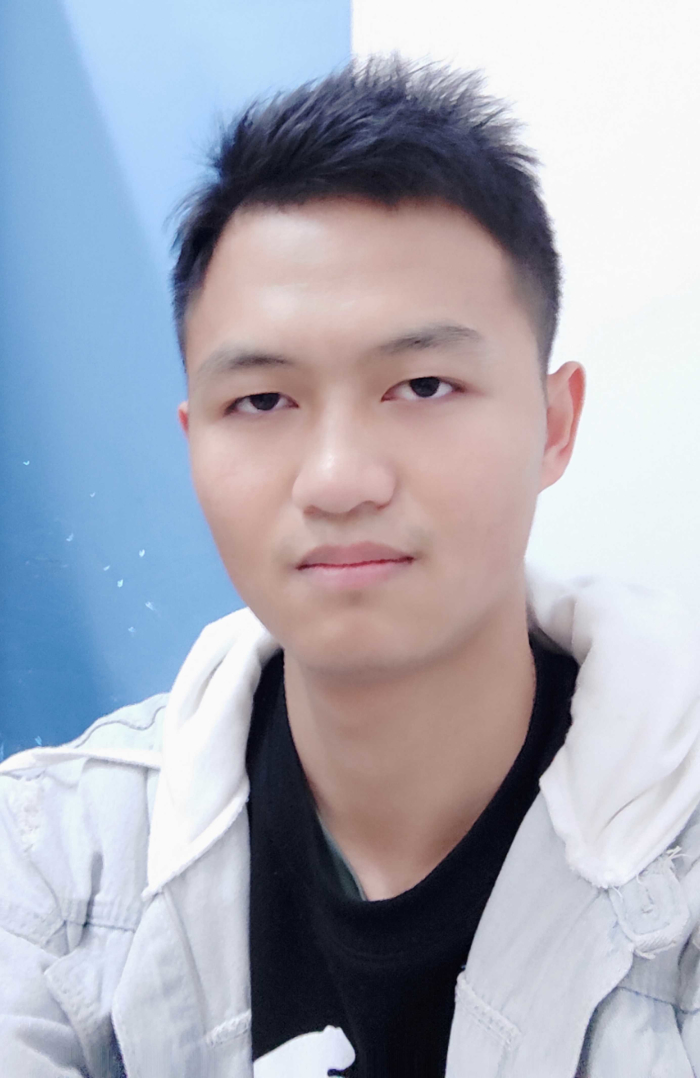}}]{Tianhao Wu} received the B.E. degree in electronic engineering from National University of Defense Technology (NUDT), Changsha, China, in 2020. He is currently pursuing the M.E. degree with the College of Electronic Science and Technology, NUDT. His research interests include infrared small target detection, light field imaging and camera calibration.
\end{IEEEbiography}

\begin{IEEEbiography}[{\includegraphics[width=1in,height=1.25in,clip,keepaspectratio]{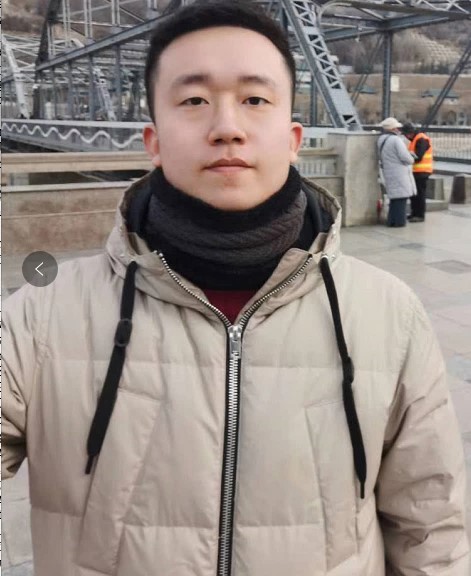}}]{Boyang Li} received the B.E. degree in Mechanical Design manufacture and Automation from the Tianjin University, China, in 2017 and M.S. degree in biomedical engineering from National Innovation Institute of Defense Technology, Academy of Military Sciences, Beijing, China, in 2020. He is currently working toward the PhD degree in information and communication engineering from National University of Defense Technology (NUDT), Changsha, China. His research interests include infrared small target detection, weakly supervised semantic segmentation and deep learning.
\end{IEEEbiography}

\begin{IEEEbiography}[{\includegraphics[width=1in,height=1.25in,clip,keepaspectratio]{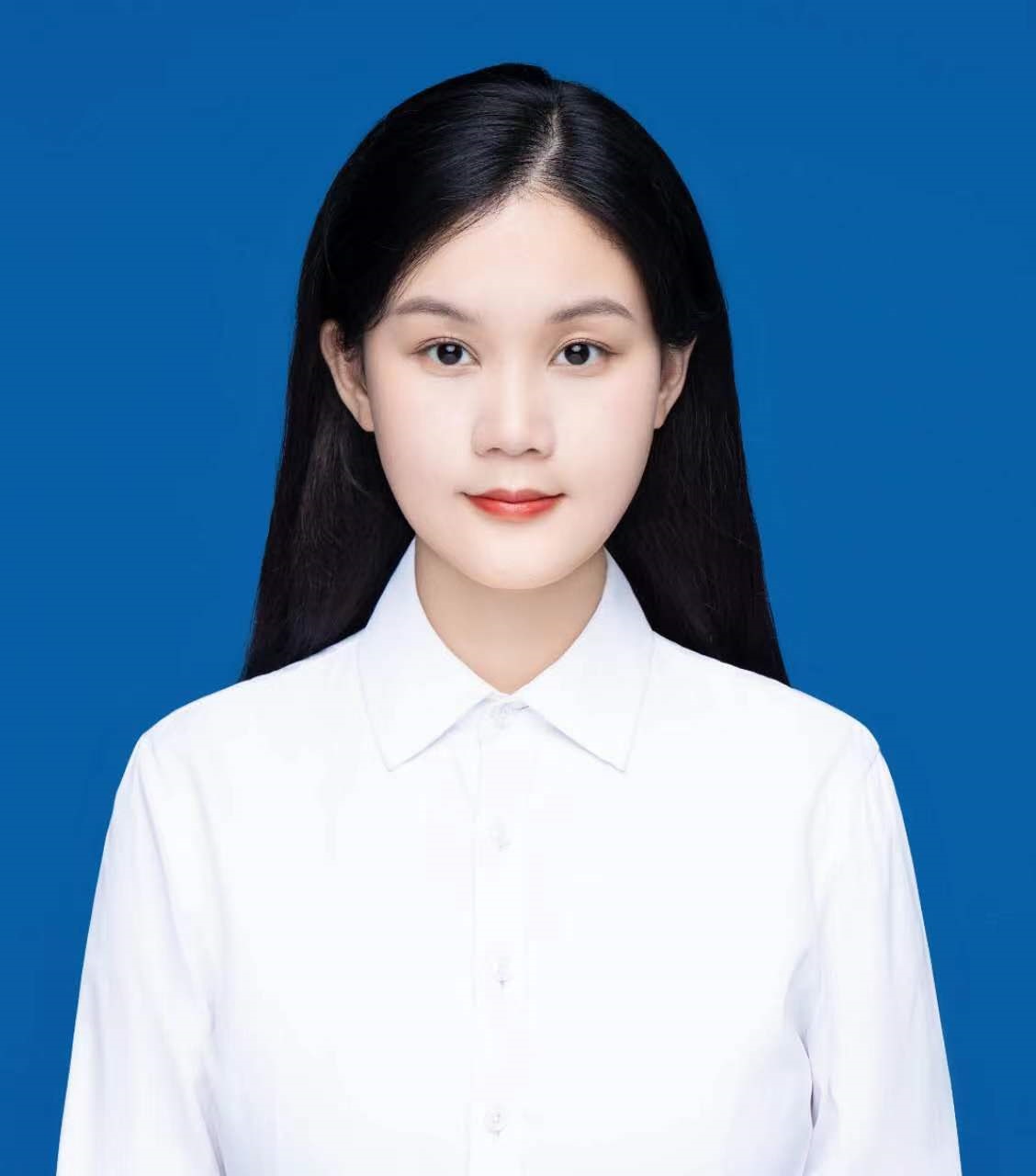}}]{Yihang Luo} received the B.E. degree in communication engineering from Hunan Normal University (NUDT), Changsha, China, in 2020. She is currently pursuing the M.E. degree with the College of Electronic Science and Technology, NUDT. Her research interests include infrared image denoising and infrared small target detection.
\end{IEEEbiography}

\begin{IEEEbiography}[{\includegraphics[width=1in,height=1.25in,clip,keepaspectratio]{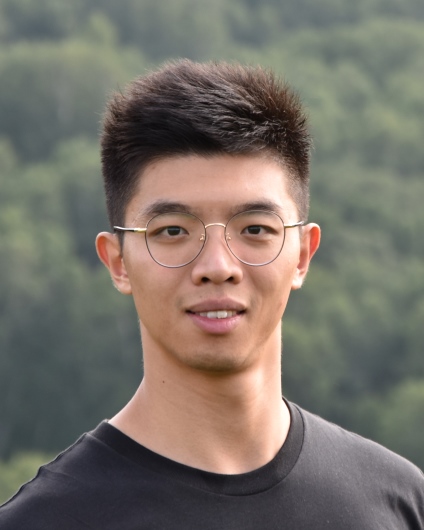}}]{Yingqian Wang} received the B.E. degree in electrical engineering from Shandong University (SDU), Jinan, China, in 2016, and the M.E. degree in information and communication engineering from National University of Defense Technology (NUDT), Changsha, China, in 2018. He is currently pursuing the Ph.D. degree with the College of Electronic Science and Technology, NUDT. His research interests focus on low-level vision, particularly on light field imaging and image super-resolution.
\end{IEEEbiography}

\begin{IEEEbiography}[{\includegraphics[width=1in,height=1.25in,clip,keepaspectratio]{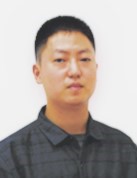}}]{Chao Xiao} received the BE degree in the communication engineering and the M.E. degree in information and communication engineering from the National University of Defense Technology (NUDT), Changsha, China in 2016 and 2018, respectively. He is currently working toward the Ph.D. degree with the College of Electronic Science in NUDT, Changsha, China. His research interests include deep learning, small object detection and multiple object tracking.
\end{IEEEbiography}

\begin{IEEEbiography}[{\includegraphics[width=1in,height=1.25in,clip,keepaspectratio]{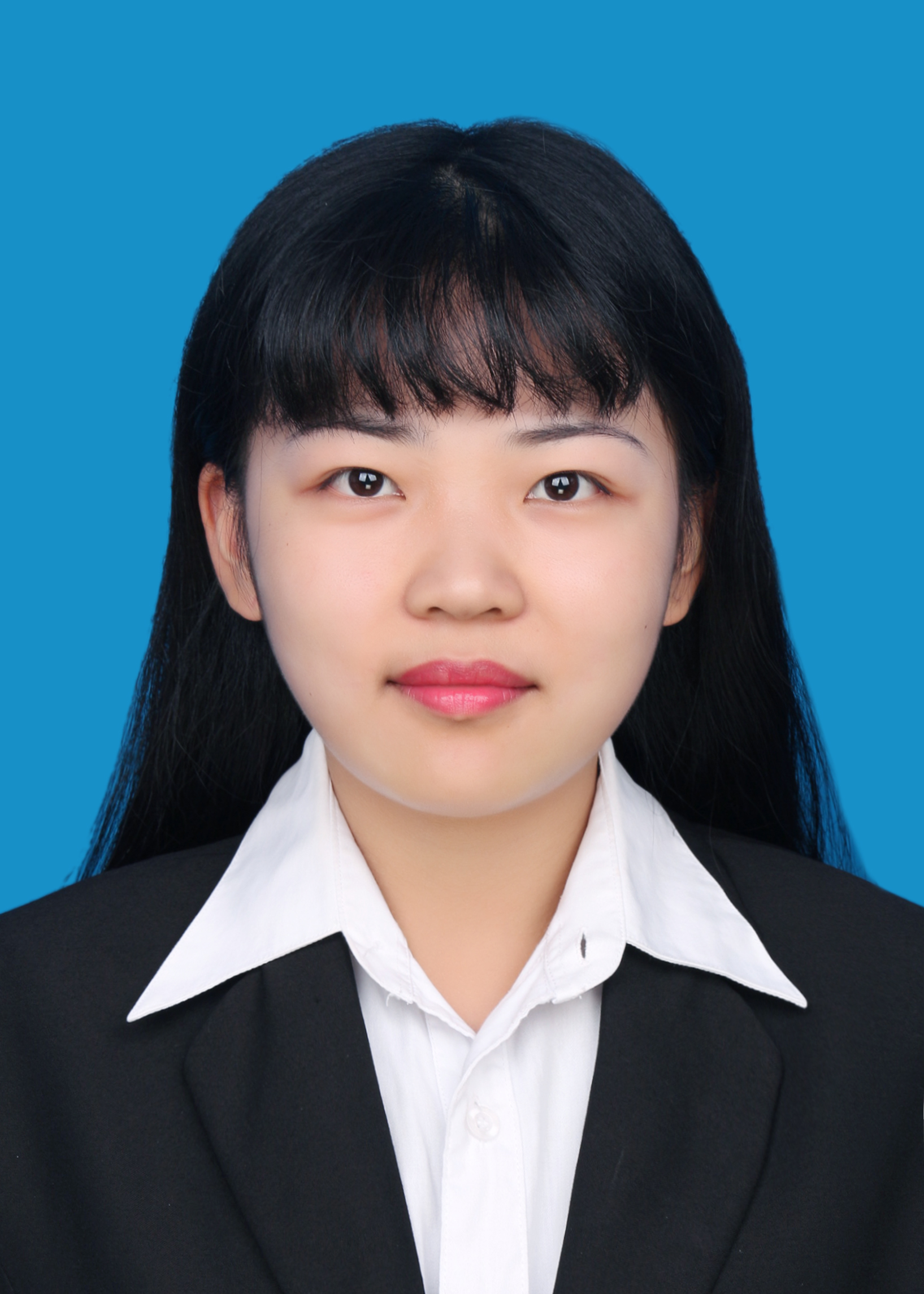}}]{Ting Liu} received the B.E. degree in electrical engineering and automation from Hunan Institute of Engineering, Xiangtan, China, in 2017, and the M.E. degree in control engineering from Xiangtan University (XTU), Xiangtan, China, in 2020. She is currently pursuing the Ph.D. degree with the College of Electronic Science in NUDT, Changsha, China. She research interests focus on signal processing, target detection and image processing.
\end{IEEEbiography}

\begin{IEEEbiography}[{\includegraphics[width=1in,height=1.25in,clip,keepaspectratio]{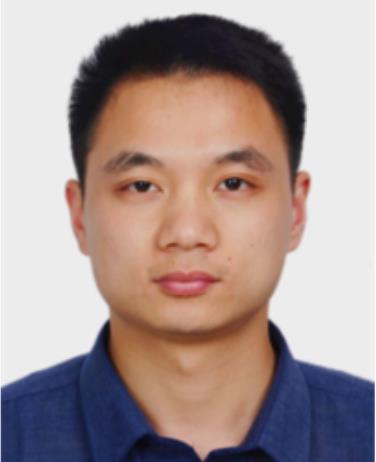}}]{Jungang Yang} received the B.E. and Ph.D. degrees from National University of Defense Technology (NUDT), in 2007 and 2013 respectively. He was a visiting Ph.D. student with the University of Edinburgh, Edinburgh from 2011 to 2012. He is currently an associate professor with the College of Electronic Science, NUDT. His research interests include computational imaging, image processing, compressive sensing and sparse representation. Dr. Yang received the New Scholar Award of Chinese Ministry of Education in 2012, the Youth Innovation Award and the Youth Outstanding Talent of NUDT in 2016.
\end{IEEEbiography}

\begin{IEEEbiography}[{\includegraphics[width=1in,height=1.25in,clip,keepaspectratio]{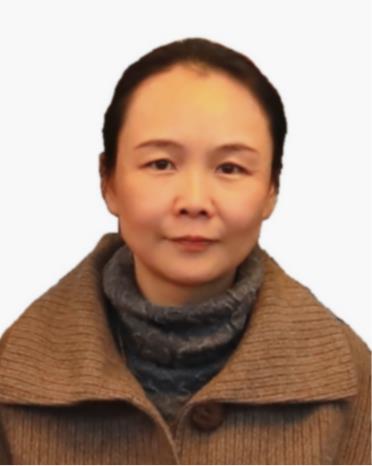}}]{Wei An} received the Ph.D. degree from the National University of Defense Technology (NUDT), Changsha, China, in 1999. She was a Senior Visiting Scholar with the University of Southampton, Southampton, U.K., in 2016. She is currently a Professor with the College of Electronic Science and Technology, NUDT. She has authored or co-authored over 100 journal and conference publications. Her current research interests include signal processing and image processing.
\end{IEEEbiography}

\begin{IEEEbiography}[{\includegraphics[width=1in,height=1.25in,clip,keepaspectratio]{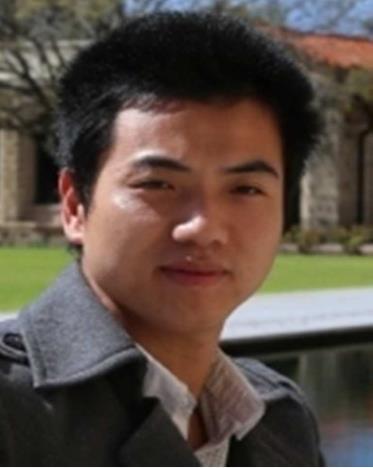}}]{Yulan Guo} received the B.E. and Ph.D. degrees from National University of Defense Technology (NUDT) in 2008 and 2015, respectively. He has authored over 100 articles at highly referred journals and conferences. His current research interests focus on 3D vision, particularly on 3D feature learning, 3D modeling, 3D object recognition, and scene understanding. He served as an associate editor for IEEE Transactions on Image Processing, IET Computer Vision, IET Image Processing, and Computers \& Graphics. He also served as an area chair for CVPR 2021, ICCV 2021, and ACM Multimedia 2021. He organized several tutorials, workshops, and challenges in prestigious conferences, such as CVPR 2016, CVPR 2019, ICCV 2021, 3DV 2021, CVPR 2022, ICPR 2022, and ECCV 2022. He is a Senior Member of IEEE and ACM.
\end{IEEEbiography}

\end{document}